%% file: main.tex
\PassOptionsToPackage{table,usenames,dvipsnames}{xcolor}
\documentclass[table]{gtech}

\usepackage[T1]{fontenc}
\usepackage[utf8]{inputenc}

\RequirePackage{tgpagella}
\RequirePackage{mathpazo}
\RequirePackage{inconsolata}
\usepackage{biolinum}

\usepackage{amsmath,amssymb,amsfonts,amsthm,bm}
\usepackage{nicefrac}
\usepackage{algorithm}
\usepackage{algpseudocode}

\usepackage{xcolor}
\usepackage[most]{tcolorbox}
\usepackage{colortbl}

\usepackage{graphicx}
\usepackage{wrapfig}
\usepackage{float}
\usepackage{subcaption}
\usepackage{caption}
\usepackage{adjustbox}
\usepackage{multicol}
\usepackage{fontawesome5}

\usepackage{array}
\usepackage{booktabs}
\usepackage{multirow}
\usepackage{bigdelim}
\usepackage{longtable}
\usepackage{tabularray}
\usepackage{tabulary}
\usepackage{makecell}
\usepackage{threeparttable}
\usepackage{tablefootnote}
\usepackage{arydshln}
\usepackage{tabularx}
\usepackage{siunitx}

\usepackage{enumitem}
\usepackage{url}
\usepackage{hyperref}
\usepackage{microtype}
\usepackage{xspace}

\usepackage{pifont}
\usepackage{bbding}
\usepackage{fontawesome5}

\usepackage{datatool}
\usepackage[normalem]{ulem} 

\theoremstyle{definition}

\newcolumntype{C}[1]{>{\centering\arraybackslash}m{#1}}

\newcommand{\ignore}[1]{}

\definecolor{CQColor}{rgb}{0.0,0.0,1.0}

\newlength\savewidth

\input{math_commands.tex}

\usepackage{titletoc}

\definecolor{pythiateal}{HTML}{0F818E}
\newcommand{\icon}{\raisebox{-6pt}{\includegraphics[width=1.5em]{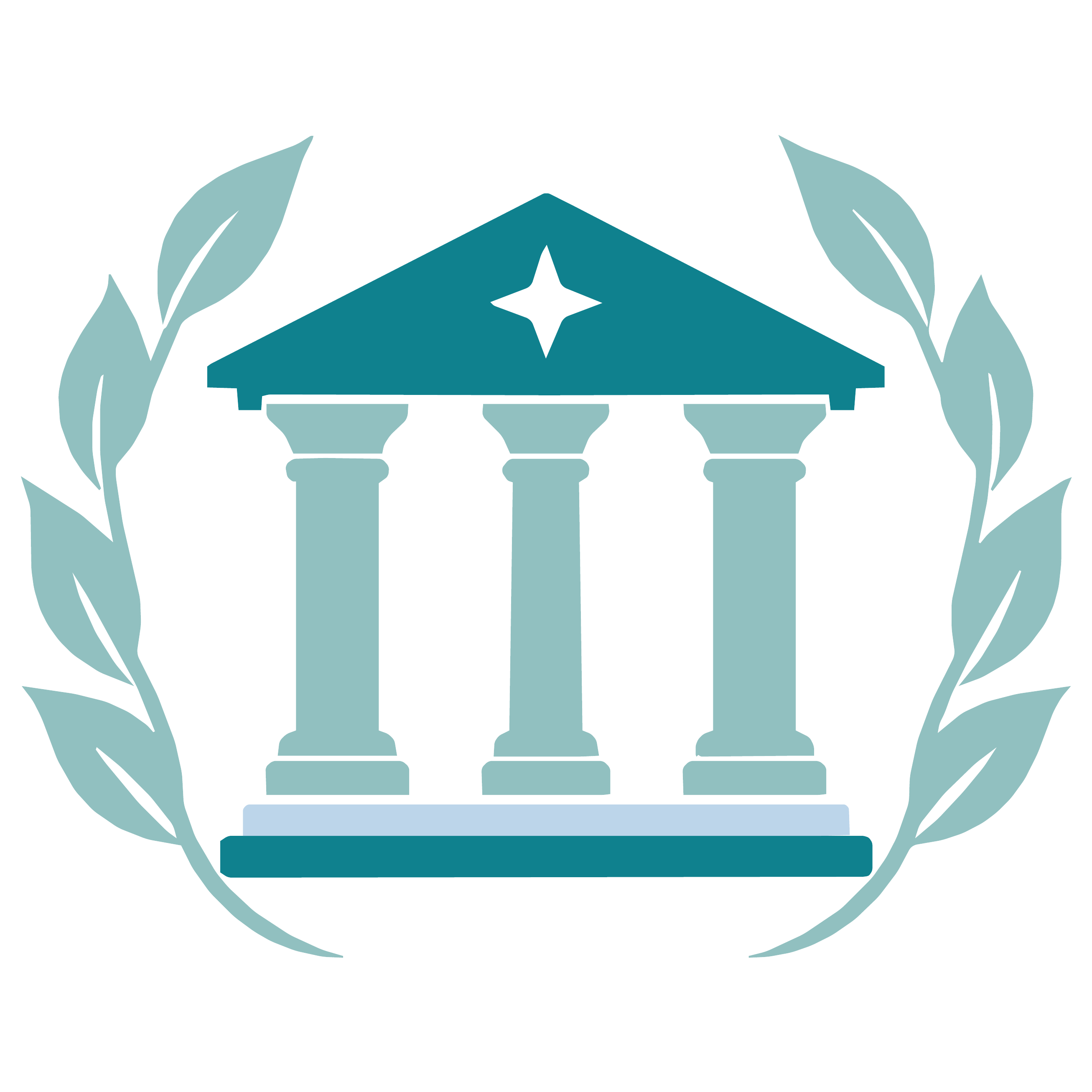}}\xspace}
\renewcommand{\title}[1]{%
  \newcommand{\titlelist}{{\huge\fontfamily{optimistic}\selectfont #1}}%
}
\title{
  \Large{Technical Report} \\[0.5em]
  \huge{\icon \textcolor{pythiateal}{Pythia}: Toward Foundation World Models for Multimodal Time Series}
}

\author[*]{Xilin Dai}
\author[*]{Hongzhou Chen}
\author[]{Yifan Hu}
\author[]{Yiding Liu}
\author[\dagger]{Zewei Dong}
\author[]{Jiang-Ming Yang}
\affiliation[]{Ant International}
\contribution[*]{Equal Contribution}
\contribution[\dagger]{Corresponding Author \\ \ \faEnvelope[regular]\ \email{\{daixilin.dxl,\ zewei.dong\}@ant-intl.com}}

\abstract{\input{sections/abstract}}
\date{October 4, 2026\vspace{-1mm}}
\gtechdata[Code]{\texttt{To be released}\vspace{-1mm}}
\gtechdata[Model]{\texttt{To be released}}

\begin{document}
\maketitle

\input{sections/introduction}
\input{sections/related_work}
\input{sections/method}
\input{sections/experiments}
\input{sections/conclusion}

\newpage
\bibliographystyle{assets/plainnat}
\bibliography{pythia}

\clearpage
\appendix
\section*{Appendix}
\titlecontents{section}[0em]
  {\vspace{0.5em}\bfseries}
  {\thecontentslabel\quad}
  {}
  {\titlerule*[0.8pc]{.}\contentspage}
\titlecontents{subsection}[1.5em]
  {\normalfont\small}
  {\thecontentslabel\quad}
  {}
  {\titlerule*[0.8pc]{.}\contentspage}
\titlecontents{subsubsection}[3em]
  {\itshape\footnotesize}
  {\thecontentslabel\quad}
  {}
  {\titlerule*[0.8pc]{.}\contentspage}
\startcontents[appendix]
\printcontents[appendix]{l}{1}{\setcounter{tocdepth}{3}}
\vspace{2em}
\input{sections/appendix}

\stopcontents[appendix]

\end{document}

%% file: math_commands.tex
\usepackage{amsmath,amsfonts,bm}

\def\eqref#1{equation~\ref{#1}}

\def\1{\bm{1}}

\DeclareMathAlphabet{\mathsfit}{\encodingdefault}{\sfdefault}{m}{sl}
\SetMathAlphabet{\mathsfit}{bold}{\encodingdefault}{\sfdefault}{bx}{n}



%% file: sections/abstract.tex
Time-series foundation models offer a unified approach to forecasting across heterogeneous domains. Textual context and auxiliary observations provide complementary information about temporal dynamics, yet reusable multimodal predictive representations remain underexplored. We introduce \textcolor{pythiateal}{\textbf{Pythia}}, a foundation world model that learns context-conditioned latent dynamics across datasets through a joint-embedding predictive architecture. A stop-gradient numerical reference guides contextual corrections to predicted future states. A separate probabilistic decoder then adapts to the frozen predictive representation and observed history, decoupling world-model pretraining from observation-space forecasting. On MUSE, Pythia-Tiny's normalized mean absolute scaled error (MASE) and weighted sum quantile loss (WSQL) are 0.6879 and 0.4269, reducing errors by \textbf{6.26\%} and \textbf{5.00\%} relative to the strongest model evaluated in the published MUSE leaderboard. Through a series of controlled experiments, we investigate how to design a time-series world model through shared pretraining and how joint-embedding predictive learning can incorporate multimodal information. The results support separating predictive representation learning from probabilistic readout and show complementary contributions from entity descriptions, events, and covariates.

%% file: sections/introduction.tex
\section{Introduction}
\label{sec:introduction}

Time-series modeling supports applications in battery state estimation, infrastructure monitoring, and physiological forecasting \citep{dai_socnet_2025,cohen2025toto,luo2026normwear2}, with forecasting methods ranging from dataset-specific full-shot Transformers such as PatchTST and GPT4TS \citep{nie2023patchtst,zhou2023gpt4ts} and linear models such as DLinear and TimePrism \citep{zeng2023dlinear,dai2026from} to recent time-series foundation models. Time-series foundation models learn shared forecasting behavior from heterogeneous collections of series, allowing one pretrained model to serve many datasets without task-specific fitting \citep{ansari2024chronos,das2024timesfm,woo2024moirai,liu2025sundial,ansari2025chronos2,liu2026falconxtimeseriesfoundation}. Numerical history, however, describes only part of the information available at a prediction origin \citep{dai2026positionuniversaltimeseries}. An entity description identifies what a signal measures; an event provides context for changes in its trajectory; and an observed covariate offers an additional view of its recent state. Multimodal foundation models extend forecasting to these complementary inputs \citep{wang2025chattime,wu2026aurora}. Paired time-series datasets and context-rich benchmarks expose both their promise and the challenge of using heterogeneous information consistently \citep{liu2024timemmd,williams2025cik,chen2026muse,xu2026fideltshighfidelitymultimodalbenchmark}.

Complementing shared pretraining, ORCA performs black-box online adaptation by learning forecast residuals conditioned on a foundation model's inputs and outputs \citep{dai2026learningcontexterrorsblackbox}. This progress raises a representation-learning question: can a model learn context-conditioned predictive states across datasets before learning how to read out future observations? Direct forecasting trains the representation and output mapping together through an observation-space loss. Separating these stages instead makes latent prediction the pretraining objective and fits the probabilistic readout to the resulting frozen representation. A joint-embedding predictive architecture (JEPA) provides a mechanism for this separation: it predicts representations of unobserved content from available observations, using an encoded target as supervision \citep{assran2023ijepa,bardes2024vjepa,assran2025vjepa2,chen2026vljepa}. For multimodal time series, the challenge is to incorporate contextual information while retaining the numerical structure needed to decode useful forecasts.

We introduce \textbf{Pythia}, a foundation world model for multimodal time series. Here, a \emph{world model} denotes a model of context-conditioned predictive states learned in latent space for observational forecasting. A numerical predictor first estimates target states from observed history. A context module then refines these estimates using entity information, events filtered by their supplied timestamps, and observed covariates. The resulting world states are matched to targets from a frozen numerical encoder. An additional objective uses a stop-gradient numerical reference to emphasize contextual correction errors relative to the prediction from numerical history. Its gradients update only the context pathway. Direction and scale alignment keep both pathways in the representation space supplied by the pretrained numerical model.

Training proceeds in two stages. The first learns latent prediction through forecasting and masked reconstruction across the pretraining corpus. The second freezes the predictive system and adapts a probabilistic decoder over observed history and predicted world states. Initialized from Chronos-2 \citep{ansari2025chronos2}, both stages learn across datasets without adaptation to individual MUSE tasks.

Pythia-Tiny provides the main benchmark result, while matched Small-model controls examine the learning recipe and contextual inputs. These experiments study how latent pretraining, numerical initialization, and multimodal integration support shared predictive representations across heterogeneous forecasting tasks.

Our contributions are threefold:
\begin{itemize}[noitemsep, topsep=0pt, leftmargin=*]
    \item \textbf{A foundation world model for time series.} We explore how joint-embedding predictive learning can produce a shared model of latent temporal dynamics through pretraining across heterogeneous datasets. A separate probabilistic readout turns its frozen predictive states into zero-shot forecasts, and controlled comparisons examine the value of this separation.
    \item \textbf{Multimodal predictive states guided by a numerical reference.} We integrate entity descriptions, timestamp-filtered events, and observed covariates as contextual corrections to numerical latent predictions. Numerical-reference guidance trains these corrections within a common representation space; component and group-level analyses reveal both their aggregate benefits and their variation across forecasting tasks.
    \item \textbf{Leading zero-shot results on MUSE.} Pythia-Tiny obtains the lowest aggregate errors among the compared methods, improving normalized mean absolute scaled error (MASE) and weighted sum quantile loss (WSQL) by \textbf{6.26\%} and \textbf{5.00\%} over the strongest published MUSE baseline, respectively. Extensive controlled experiments connect these forecasting results to the design of the predictive representation, multimodal integration, and two-stage training.
\end{itemize}

%% file: sections/related_work.tex
\section{Related Work}
\label{sec:related}

\subsection{Time-series foundation models}
Large-scale pretraining on heterogeneous series produces reusable forecasting priors, as demonstrated by Chronos, TimesFM, and Moirai \citep{ansari2024chronos,das2024timesfm,woo2024moirai}. Subsequent work expands the architectures and objectives used for shared forecasting: Time-MoE explores sparse experts, while Sundial learns continuous predictive distributions through flow matching \citep{shi2025timemoe,liu2025sundial}. Chronos-2 supports univariate, multivariate, and covariate-informed prediction \citep{ansari2025chronos2}. Beyond numerical inputs, ChatTime combines time series and text in a shared model, and Aurora studies universal generative multimodal forecasting \citep{wang2025chattime,wu2026aurora}. Pythia makes context-conditioned latent prediction its pretraining objective and separately adapts a probabilistic decoder to its frozen predictive states.

\subsection{Toward multimodal time-series forecasting}
Textual and auxiliary observations provide information beyond numerical history. Time-MMD pairs numerical and textual data, Context is Key (CiK) isolates problems requiring textual information, and MUSE evaluates heterogeneous contextual inputs under a shared benchmark \citep{liu2024timemmd,williams2025cik,chen2026muse}. GPT4MTS and Texts as Time Series integrate text with numerical prediction \citep{jia2024gpt4mts,li2026tats}; REST-TS addresses text collapse through forecast-residual supervision \citep{nguyen2026restts}. Generative artificial intelligence (AI) methods LLMTime and Time-LLM use large language models (LLMs) via numerical serialization and reprogramming, respectively \citep{gruver2023llmtime,jin2024timellm}. Pythia uses semantic context to refine a numerical latent estimate, with a detached numerical reference guiding the correction before readout training. We evaluate on MUSE using its published baselines. Appendix~\ref{app:related_work} provides the extended discussion, including predictive representation learning.

%% file: sections/method.tex
\section{Pythia}
\label{sec:method}

\begin{figure}[t]
\centering
\includegraphics[width=\linewidth]{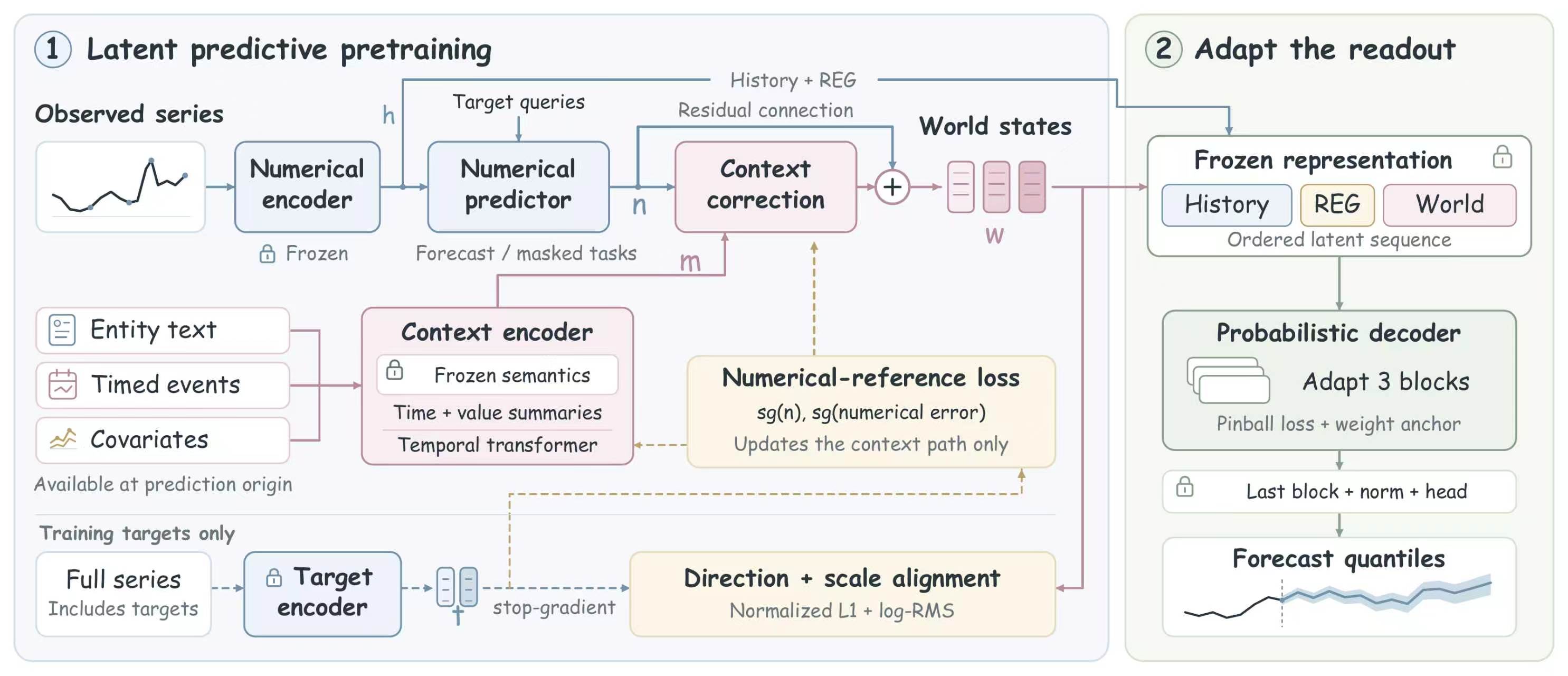}
\caption{Pythia's default two-stage recipe. Stage 1 learns numerical predictions $n$ from memory $h$ and a residual correction using context memory $m$, aligning world states $w$ with stop-gradient ($\operatorname{sg}$) target states $t$ encoded from full numerical sequences. Matching uses absolute-distance ($L_1$) and log root-mean-square (RMS) terms. A detached numerical reference guides only the context pathway. Stage 2 freezes the predictive representation and adapts three decoder blocks over ordered history, the special REG token, and world states using pinball loss and a parameter anchor; the final decoder block, normalization, and quantile head remain frozen. Targets provide supervision only.}
\label{fig:architecture}
\end{figure}

\subsection{Forecasting through predictive latent states}

For prediction origin $T$ and forecast horizon $H$, let $x_{1:T}$ denote the observed numerical history and $c_T$ the accompanying context retained by input preprocessing. The forecasting task is to estimate future observations $x_{T+1:T+H}$, including predictive quantiles, with a shared model across datasets. Pythia represents these future observations through predicted latent states $w$, first learning the mapping $(x_{1:T},c_T)\mapsto w$ and then fitting a probabilistic readout to the frozen mapping. Figure~\ref{fig:architecture} summarizes this division of learning objectives.

We divide numerical sequences into patches of 16 observations. The first eight blocks of a pretrained Chronos-2 model form a frozen numerical encoder $E$. For a sampled task, $E$ encodes the visible numerical view into a memory $h$. A separate reference pass encodes the complete numerical view, including the target patches, with the same normalization statistics. Its target states $t_j$, indexed by target patch $j$, are stop-gradient references and are never inputs to the predictor. During forecasting, future numerical values are unobserved and excluded from the predictor's numerical memory.

A trainable predictor $P_\theta$, with parameters $\theta$, maps learned target queries $q_j$ and visible memory to numerical predictions $n_j=P_\theta(q_j,h)$. Each predictor block contains query self-attention, cross-attention to numerical memory, and a feed-forward network. Queries combine a mask embedding with position, calendar features, frequency, signed logarithmic lead, and a task identifier. Pretraining mixes causal forecasting and contiguous masked-block reconstruction with equal probability, supervising at most six target patches per sampled task. The task embedding distinguishes prediction beyond an observed prefix from reconstruction within a partially observed sequence.

\subsection{Context as a residual correction}

Parameters $\psi$ belong to the context encoder and correction module. The context encoder $C_\psi$ receives entity information, event content with timestamps, and named numerical covariates. A frozen pretrained multimodal encoder supplies cached semantic embeddings. Event content and time projections are added to form one token per event; similarly, a covariate-name embedding and a numerical-summary projection are added to form one token per covariate. Numerical summaries describe observed values through level, variation, trend, coverage, and coarse temporal bins. A temporal transformer combines these tokens into context memory $m=C_\psi(c_T)$. Events are retained only when their supplied timestamps are no later than the prediction origin, and covariate summaries use historical observations only.

Using residual map $g_\psi$, learned output projection $W_\Delta$, layer normalization $\operatorname{LN}$, and multi-head cross-attention $\operatorname{Attn}$ (query, key, value), we compute the world state
\begin{equation}
 w_j=g_\psi(n_j,m)
 =n_j+W_\Delta\,\operatorname{Attn}\!\left(\operatorname{LN}(n_j),m,m\right).
 \label{eq:correction}
\end{equation}
The residual formulation keeps the numerical estimate available while allowing complementary context to change the predicted state. When context is unavailable, the correction is masked and the numerical estimate is retained. The numerical estimate also serves as the reference for contextual refinement below; it has no separate first-stage forecasting loss.

\subsection{Latent supervision and numerical-reference guidance}

For predicted and target latent vectors $z,t\in\mathbb R^d$, $d$ is the width and $z_k$ the $k$th coordinate. Write $\operatorname{sg}$ for stop-gradient and $\|\cdot\|_1,\|\cdot\|_2$ for the absolute-sum and Euclidean norms. With $\epsilon_n=10^{-12}$, unit normalization is $u(z)=z/\max(\|z\|_2,\epsilon_n)$. Brackets $\langle\cdot\rangle$ average supervised patches within a sample and then samples, incorporating sampling weights. Normalized patch error $e$ and JEPA loss are
\begin{equation}
 e(z,t)=\frac{1}{d}\big\|u(z)-u(\operatorname{sg}(t))\big\|_1,
 \qquad
 \mathcal L_{\mathrm{JEPA}}=\left\langle e(w_j,t_j)\right\rangle.
 \label{eq:jepa}
\end{equation}
Normalization constrains latent direction but discards magnitude. We also align log-RMS magnitude $r$, with $\epsilon=10^{-8}$ and unit-threshold smooth $L_1$ penalty $\operatorname{SmoothL1}$:
\begin{equation}
 r(z)=\log\sqrt{\frac{1}{d}\sum_{k=1}^{d}z_k^2+\epsilon},
 \qquad
 \mathcal L_{\mathrm{scale}}
 =\left\langle\operatorname{SmoothL1}\!\left(r(w_j),r(\operatorname{sg}(t_j))\right)\right\rangle,
 \label{eq:scale}
\end{equation}
Together, these objectives align the direction and magnitude of the world states to the frozen numerical representation used by the decoder.

To guide context correction relative to the numerical estimate, we recompute the corrected state with a detached numerical input. Define the corrected error
$a_j=e(g_\psi(\operatorname{sg}(n_j),m),t_j)$ and detached numerical error
$b_j=\operatorname{sg}(e(n_j,t_j))$.
For corrected and numerical patch errors $a,b$, define the penalty $\ell(a,b)$ and the reference objective as
\begin{equation}
 \ell(a,b)=
 \begin{cases}
 b\log\!\left(1+\exp\!\left((a-b)/b\right)\right),&b>0,\\
 a,&b=0,
 \end{cases}
 \qquad
 \mathcal L_{\mathrm{ref}}=\left\langle\ell(a_j,b_j)\right\rangle.
 \label{eq:reference}
\end{equation}
We evaluate Equation~\ref{eq:reference} in a numerically stable form given in Appendix~\ref{app:implementation}. The numerical error provides both a reference level and a patch-specific positive scale. For $b>0$, the $a$-derivative is the logistic sigmoid $\operatorname{sigmoid}((a-b)/b)$: larger correction errors relative to the numerical reference receive greater gradient emphasis, with a smooth transition around that reference. This guidance conditions the context gradient on numerical error; it does not enforce $a\leq b$. Detaching $n_j$ and $b_j$ confines this loss to the context encoder and correction module. The world-state matching and scale losses update both the numerical predictor and the context pathway. The full first-stage objective is
\begin{equation}
 \mathcal L_1=\mathcal L_{\mathrm{JEPA}}+0.05\mathcal L_{\mathrm{scale}}+\mathcal L_{\mathrm{ref}}.
 \label{eq:stage1}
\end{equation}
Numerical-reference guidance supplies context-only gradient emphasis without requiring negative context examples. The target encoder is maintained as an exponential moving average (EMA) copy of the numerical encoder; because the latter is frozen in the default recipe, the target encoder remains fixed. Supervision acts on latent states rather than decoded future values.

\subsection{Independent probabilistic readout}

After latent pretraining, we freeze the numerical encoder, predictor, context encoder, and correction module. For observed-history memory $h_{\mathrm{history}}$, the backbone's special REG token state $h_{\mathrm{REG}}$, and predicted future world states $w_{\mathrm{future}}$, the decoder receives $[h_{\mathrm{history}},h_{\mathrm{REG}},w_{\mathrm{future}}]$ in this order. Observed history and predicted world states jointly support the readout. The second-stage loss learns this combination while preserving the pretrained representation.

The readout is initialized from the remaining Chronos-2 blocks and its native quantile head. We adapt three decoder blocks while keeping the last block, final normalization, and quantile head frozen. For future observation $x$, predicted quantile $\hat x^{(\tau)}$ at native probability $\tau$, residual $v=x-\hat x^{(\tau)}$, and indicator function $\mathbf 1\{\cdot\}$, the pinball loss is $\rho_\tau(v)=(\tau-\mathbf 1\{v<0\})v$. Its average over quantiles and observed future targets is $\mathcal L_{\mathrm{pinball}}$. The parameter-anchor penalty $\mathcal L_{\mathrm{anchor}}$ regularizes displacement from the pretrained decoder, giving the second-stage objective
$\mathcal L_2=\mathcal L_{\mathrm{pinball}}+10^{-3}\mathcal L_{\mathrm{anchor}}$.
Appendix~\ref{app:implementation} specifies the adaptation and optimization details. This stage uses the pretraining corpus rather than task-specific MUSE data. At inference, the same frozen predictive system and adapted decoder serve all benchmark datasets without additional fitting.

%% file: sections/experiments.tex
\section{Experiments}
\label{sec:experiments}

\subsection{Experimental setup}

\paragraph{Benchmark and metrics.}
We evaluate on MUSE \citep{chen2026muse}, a multimodal forecasting benchmark with 14 datasets spanning eight domains and six observation frequencies. Evaluation covers 12,729 test windows and 29 dataset--frequency--horizon scoring groups. We report MASE for median forecasts and WSQL over nine predictive deciles. Following MUSE, each group is normalized against Seasonal Naive on matching coverage, and the overall score is the equal-group geometric mean. Lower scores are better. Appendix~\ref{app:protocol} gives the evaluation protocol.

\paragraph{Unified pretraining and evaluation.}
Both Pythia stages train across heterogeneous time series, without dataset-specific adaptation on MUSE. We reserve a held-out subset of the pretraining corpus for checkpoint selection and report forecasting results on the MUSE test set. Tiny is our headline configuration; Medium provides a larger representative, and Small is the common platform for matched mechanism experiments. Main results use the selected readout checkpoint. Capacity, budget, and batch analyses instead compare final checkpoints at the declared budgets, keeping training-horizon effects distinct from checkpoint selection.

\paragraph{Comparisons.}
We retain all 31 baseline results published in MUSE, with their original zero-shot or dataset-specific training protocols. These cover statistical forecasters, numerical foundation models, multimodal models, and LLM-based approaches, including DLinear \citep{zeng2023dlinear}, PatchTST \citep{nie2023patchtst}, TimesNet \citep{wu2023timesnet}, and TimeMixer \citep{wang2024timemixer}. Chronos2 has the strongest published aggregate scores and supplies Pythia's numerical initialization. The benchmark comparison evaluates the complete Pythia system, including its additional pretraining; matched Small experiments examine the contributions of predictive pretraining, context, and readout adaptation.

\subsection{Leading zero-shot performance on MUSE}
\input{sections/main_table}

\begin{figure}[t]
\centering
\includegraphics[width=\linewidth]{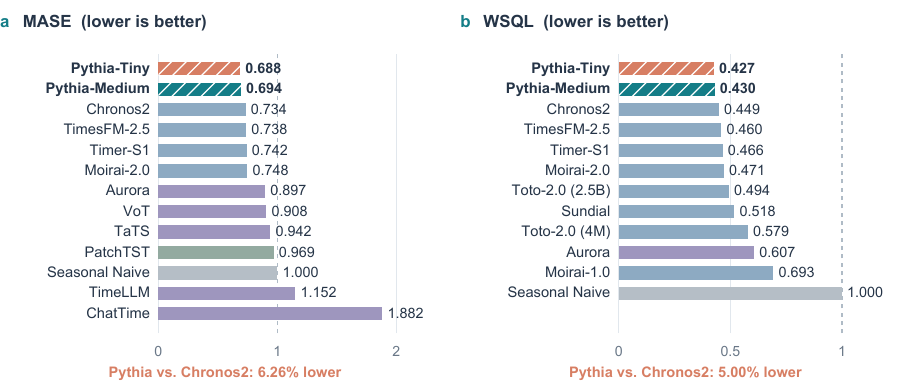}
\caption{MUSE forecasting with Pythia-Tiny and Medium. The MASE panel includes leading numerical foundation models and representative multimodal, LLM-based, dataset-specific, and statistical methods. The WSQL panel includes all published baselines with a reported probabilistic score; the two panels therefore have different rosters. Bars start at zero, dashed lines mark Seasonal Naive, and hatching identifies Pythia. All 31 published baselines remain in Table~\ref{tab:main}.}
\label{fig:main-results}
\end{figure}

\paragraph{Overall performance.}
Pythia-Tiny achieves \textbf{0.6879 MASE} and \textbf{0.4269 WSQL}, improving on Chronos2 (0.7338/0.4493) by \textbf{6.26\%} and \textbf{5.00\%}, respectively (Table~\ref{tab:main}; Figure~\ref{fig:main-results}). It also outperforms the other leading numerical foundation models, including TimesFM-2.5, Timer-S1, and Moirai-2.0. Relative to the multimodal foundation model Aurora (0.8973/0.6071), Tiny reduces the two errors by 23.34\%/29.69\%. Pythia thus improves on the strongest published MUSE results with one shared model across the benchmark. Appendix~\ref{app:quantile-diagnostics} complements WSQL with empirical quantile coverage and ordering diagnostics.

The gain is not confined to Tiny: Medium obtains 0.6943/0.4300, Small 0.6981/0.4373, and Large 0.6955/0.4279. Their training budgets differ; we examine capacity at matched budgets below.

\paragraph{Distribution across tasks.}
Tiny improves over Seasonal Naive in 27 of 29 groups for MASE and 28 of 29 for WSQL. Its short-, medium-, and long-horizon MASE scores are 0.6695, 0.7228, and 0.6721; WSQL is 0.4960, 0.4385, and 0.2657. The gains are therefore distributed across the benchmark, with some task-specific exceptions. Tiny's MASE is slightly above 1 on short-horizon ACL18 stock and TimeCAP finance. No family member dominates every group: Medium, for example, improves on Tiny at weekly and monthly frequencies. Appendix Figure~\ref{fig:groups} and Table~\ref{tab:groups} retain all family-level slices and individual group scores, using one checkpoint per model throughout.

\subsection{What makes predictive world states useful?}
\begin{figure}[t]
\centering
\includegraphics[width=\linewidth]{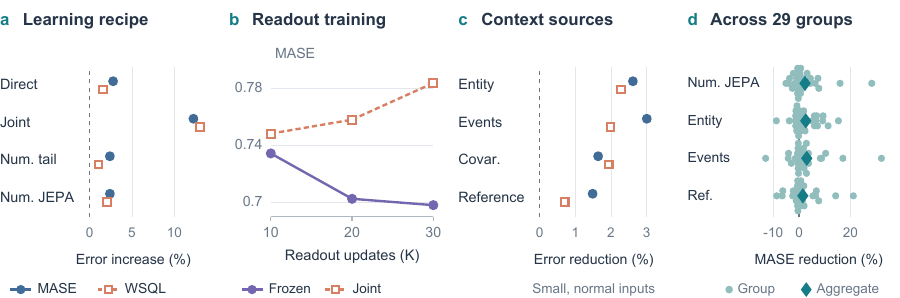}
\caption{Predictive pretraining and multimodal context, using Small for matched comparisons. Update counts use K for thousands. (a) Error increases over Pythia under equal 60K total updates: direct forecasting, joint readout training, numerical decoder adaptation (Num.\ tail), and numerical-only two-stage learning (Num.\ JEPA). (b) MUSE MASE during readout training after the same Stage~1, with the representation frozen or jointly updated. (c) Error reductions of the full model over entity, event, covariate (Covar.), and reference (Ref.) removal controls. (d) Group-level MASE reductions over the numerical-only and three source/reference controls; diamonds show aggregate reductions. Negative group effects remain visible. Exact scores and both-metric group results appear in Tables~\ref{tab:ablation} and~\ref{tab:numerical-groups} and Figure~\ref{fig:context-groups}.}
\label{fig:evidence}
\end{figure}

\paragraph{Separate representation learning from readout adaptation.}
Figure~\ref{fig:evidence}a compares five learning recipes at global batch 1,024 and 60K total updates. Pythia uses 30K latent-pretraining updates followed by 30K readout updates. Direct forecasting trains through the observation-space objective for 60K updates. Joint readout adaptation reuses Pythia's Stage~1 but continues updating its representation in Stage~2. Numerical decoder adaptation trains native Chronos decoder blocks for 60K updates. Numerical-only two-stage learning preserves Pythia's task mixture, latent objective, and 30K+30K schedule, but disables context and reference guidance throughout training. Equal updates do not imply equal computation.

Pythia-Small reduces MASE/WSQL by 2.70\%/1.55\% over direct forecasting and 10.91\%/11.51\% over joint readout adaptation. Numerical decoder adaptation obtains 0.7148/0.4419 and numerical-only two-stage learning 0.7148/0.4464, compared with Pythia's 0.6981/0.4373. The latter comparison gives 2.34\%/2.03\% reductions from incorporating multimodal context and its reference-guided objective within the same two-stage recipe. The shared-Stage~1 trajectories clarify the joint comparison: between 10K and 30K readout updates, MASE falls from 0.7343 to 0.6981 with the representation frozen, but rises from 0.7481 to 0.7836 when it is jointly updated (Figure~\ref{fig:evidence}b). WSQL follows the same divergence. These results support separating predictive representation learning from observation-space readout adaptation.

\paragraph{Preserve information needed by the observation-space interface.}
The latent and readout ablations reveal complementary requirements (Appendix Figure~\ref{fig:latent-readout}). Removing latent magnitude alignment raises MASE/WSQL by 12.38\%/14.38\%, while removing the task embedding raises them by 4.48\%/5.39\%. Direction-only latent matching is therefore insufficient for the tested readout, and the forecasting/reconstruction distinction is useful. The task mixture gives the best MASE among the task variants, although masked-only pretraining has lower WSQL. On the readout side, removing observed history raises errors by 61.09\%/66.36\%. The readout combines observed numerical memory with the predictive information in world states. A separately trained query decoder also benefits from world states, but remains weaker than the adapted native readout, highlighting the importance of a compatible representation--decoder interface.

\subsection{How multimodal information contributes}

Context is heterogeneous across MUSE windows: entity descriptions are present in all 12,729 instances, events in 73.96\%, and historical covariates in 82.53\%; 59.92\% contain all three. These counts use the model's effective input masks. The shared model operates across these different combinations of context. Appendix~\ref{app:context-coverage} reports the combinations and preprocessing rules.

The source-removal controls in Figure~\ref{fig:evidence}c keep the Small training protocol fixed. Full Pythia reduces MASE by 2.61\%, 3.00\%, and 1.64\% relative to the no-entity, no-event, and no-covariate controls, respectively; all three also improve WSQL. Without reference guidance, adding all context to the numerical-only two-stage model reduces MASE/WSQL by 0.87\%/1.34\%. Adding reference guidance with the same contextual inputs yields a further 1.49\%/0.70\% reduction. These separate comparisons support contributions from both contextual information and its training objective; Appendix Table~\ref{tab:context-reference} reports the intermediate context combinations. At the fixed Small checkpoint, contextual correction lowers mean latent error in all 29 groups, reducing the group-macro error from 0.03614 to 0.02291 (Appendix~\ref{app:latent-diagnostics}).

The paired numerical-only comparison improves 17 of 29 groups in MASE and 18 in WSQL, with both metrics improving on 15 groups (Figure~\ref{fig:evidence}d). Benefits are heterogeneous: short-horizon CiK transport and medium-horizon daily retail show large gains, whereas several finance and healthcare groups favor numerical-only learning. Source-removal controls likewise show both positive and negative group effects. Further interface controls show that removing semantic-embedding normalization or replacing paired tokens with separate content/attribute tokens worsens both aggregate metrics (Appendix~\ref{app:additional}). Appendix~\ref{app:context} further resolves the fixed-checkpoint null/random responses by scoring group and context availability.

\subsection{How should a numerical foundation model initialize Pythia?}
\begin{figure}[t]
\centering
\includegraphics[width=\linewidth]{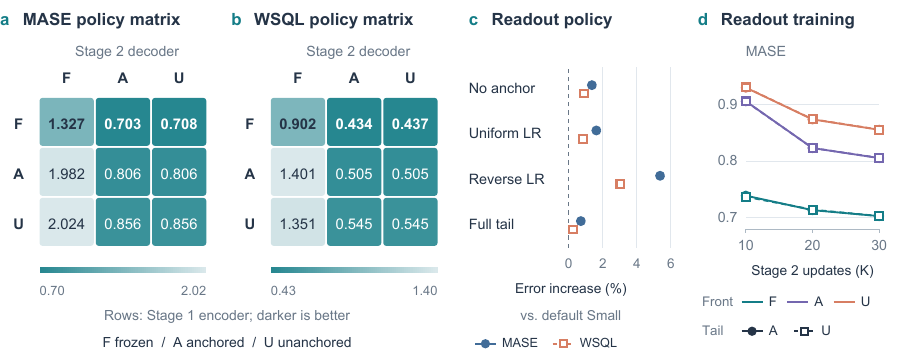}
\caption{Transferring Chronos-2 into a multimodal world model. (a,b) All nine Small encoder/decoder policies: frozen (F), anchored adaptation (A), and unanchored adaptation (U). Rows control the numerical encoder in Stage~1; columns control the decoder in Stage~2. Darker cells indicate lower errors within each metric. F decoder columns use zero readout updates; A/U columns use validation-selected checkpoints. (c) Changes to the default Small decoder recipe; LR denotes learning rate. (d) All six adapted-decoder trajectories: color denotes the Stage~1 encoder policy, solid circles/dashed squares the Stage~2 anchor policy. Paired A/U curves nearly coincide. All encoders are frozen in Stage~2. F/U selects 25K, while (d) retains its 30K endpoint; full scores and policy definitions appear in Appendix~\ref{app:initialization}.}
\label{fig:numerical-initialization}
\end{figure}

Chronos-2 supplies both the numerical representation and its native probabilistic interface. Their roles need not imply the same adaptation policy. We cross frozen, anchored, and unanchored encoder adaptation during Stage~1 with the corresponding decoder policies during Stage~2 (Figure~\ref{fig:numerical-initialization}a,b). All nine configurations share pretrained initialization; adapted blocks use the same learning-rate multiplier. The representation is frozen throughout Stage~2, including after Stage~1 encoder adaptation.

Preserving the encoder gives the lowest errors under every decoder policy, whereas the decoder benefits substantially from adaptation: F/F obtains 1.3271/0.9020 MASE/WSQL, compared with 0.7031/0.4337 for F/A. With adapted decoders, anchoring the encoder improves over unanchored adaptation, but neither reaches the frozen-encoder result. These comparisons favor a stable numerical representation with an adapted observation-space interface; they do not isolate encoder drift from changes in target dynamics or optimization.

Anchoring is also policy-dependent. At matched 30K endpoints, anchored and unanchored decoder curves nearly coincide (Figure~\ref{fig:numerical-initialization}d). Yet removing the anchor from the default descending-rate decoder raises MASE/WSQL by 1.36\%/0.88\%. Reversing its layerwise learning rates is more damaging than expanding adaptation to the entire tail (Figure~\ref{fig:numerical-initialization}c). The experiments support selective reuse of the numerical foundation model, with distinct choices for preserving its latent space and adapting its readout. Appendix~\ref{app:initialization} analyzes every policy and related transfer control.

\subsection{Capacity and training budget}
\begin{figure}[t]
\centering
\includegraphics[width=\linewidth]{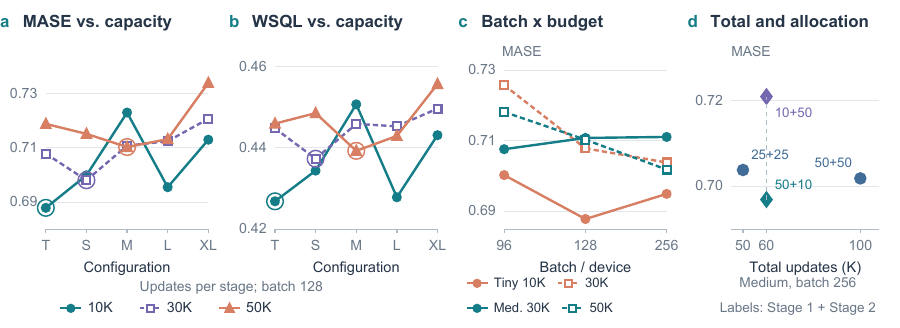}
\caption{Interacting training choices at final checkpoints. (a,b) The complete five-size, three-budget grid at batch 128 per device; circles mark each budget's minimum. T/S/M/L/XL denote Tiny/Small/Medium/Large/XLarge. (c) Batch effects on Tiny and Medium (Med.); solid/dashed lines use 10K/30K per stage for Tiny and 30K/50K for Medium. (d) Medium at batch 256 with different total updates and Stage~1+Stage~2 allocations; diamonds share 60K total updates. Each declared budget has its own training schedule.}
\label{fig:scaling}
\end{figure}

The full grid reveals a local capacity--budget relationship (Figure~\ref{fig:scaling}a,b): the best configuration on both metrics shifts from Tiny to Small to Medium as the per-stage budget increases from 10K to 30K to 50K. The corresponding optimum does not improve monotonically. Medium benefits from the larger budgets, whereas Tiny and XLarge do not.

Batch effects also depend on the training horizon (Figure~\ref{fig:scaling}c). Tiny prefers batch 128 at 10K, but improves with batch 96$\rightarrow$128$\rightarrow$256 at 30K; Medium's smallest batch is best at 30K, whereas its largest is best at 50K. Stage allocation adds another interaction: at 60K total updates and batch 256, Medium's 50K+10K recipe improves MASE/WSQL by 3.34\%/2.46\% over 10K+50K (Figure~\ref{fig:scaling}d). Even equal sampled exposure need not give equal performance when batch and update count differ.

Appendix~\ref{app:budget} reports the full batch slices, stage allocations, equal-exposure comparisons, and within-run trajectories. These experiments vary capacity and training exposure over a fixed corpus; they do not measure scaling with independent data volume.

%% file: sections/main_table.tex
\begin{table}[t]
\centering
\small
\caption{Main MUSE results, retaining all 31 baselines from MUSE Table 3 \citep{chen2026muse} and their published zero-shot or dataset-specific training protocols. Scores are Seasonal-Naive-normalized geometric means; lower is better ($\downarrow$). Model-size suffixes m/M and B denote millions and billions of parameters. Baselines are ordered by MASE down the left block, then the right. A dash denotes an unreported probabilistic score. Pythia configurations have different training budgets; their controlled capacity comparison is in Figure~\ref{fig:scaling}.}
\label{tab:main}
\begin{tabular*}{\linewidth}{@{\extracolsep{\fill}}lrr@{\hspace{1.2em}}lrr@{}}
\toprule
Model & MASE & WSQL & Model & MASE & WSQL\\
\midrule
Chronos2 & 0.7338 & 0.4493 & TimeMixer & 1.0162 & --\\
TimesFM-2.5-200m & 0.7381 & 0.4601 & CALF & 1.0484 & --\\
Timer-S1 & 0.7419 & 0.4661 & Moirai-1.0-R-large & 1.0559 & 0.6931\\
Moirai-2.0-R-small & 0.7477 & 0.4715 & Timer-base-84m & 1.0905 & --\\
Sundial-base-128m & 0.7788 & 0.5184 & TimeCMA & 1.1345 & --\\
Toto-2.0-2.5B & 0.7913 & 0.4941 & TimeLLM & 1.1521 & --\\
Aurora & 0.8973 & 0.6071 & AutoTimes & 1.3201 & --\\
VoT & 0.9079 & -- & MovingAverageSeasonal & 1.3740 & --\\
Toto-2.0-4m & 0.9207 & 0.5785 & LastValueNaive & 1.4506 & --\\
TaTS & 0.9418 & -- & ChatTime-1-7B-Base & 1.8817 & --\\
DLinear & 0.9478 & -- & MeanForecast & 2.2575 & --\\
TimesNet & 0.9541 & -- & TimeMoE-200M & 2.3387 & --\\
TESS & 0.9543 & -- & ZeroForecast & 4.5881 & --\\
PatchTST & 0.9686 & -- & Qwen3.6-27B-Prompt & 8.5229 & --\\
GPT4MTS & 0.9726 & -- & Gemma-4-26B-A4B-it & 9.9501 & --\\
SeasonalNaive & 1.0000 & 1.0000 &  &  & \\
\midrule
\multicolumn{6}{l}{\emph{Pythia family: shared pretraining, zero-shot MUSE evaluation}}\\
\textbf{Pythia-Tiny} & \textbf{0.6879} & \textbf{0.4269} & Pythia-Small & 0.6981 & 0.4373\\
Pythia-Medium & 0.6943 & 0.4300 & Pythia-Large & 0.6955 & 0.4279\\
\bottomrule
\end{tabular*}
\end{table}

%% file: sections/conclusion.tex
\section{Conclusion}
\label{sec:conclusion}

Pythia explores foundation world modeling for time series through shared latent predictive pretraining and a separately adapted probabilistic readout. Numerical-reference guidance helps entity information, events, and covariates refine numerical predictions. Pythia-Tiny leads the MUSE comparison, reducing normalized MASE and WSQL by 6.26\% and 5.00\% relative to the strongest model evaluated in the published MUSE leaderboard. Matched Small-model controls support separating predictive pretraining from readout adaptation; removing contextual inputs worsens aggregate accuracy, although their benefits vary across tasks.

Together, these experiments characterize how numerical initialization, contextual correction, and stage-specific adaptation shape multimodal predictive learning. The results support a shared latent interface between contextual pretraining and probabilistic forecasting, with the benefit of individual design choices assessed under matched training conditions.

%% file: sections/appendix.tex
\section{Implementation and training details}
\label{app:implementation}

\subsection{Model components and configurations}
Pythia reuses the first eight blocks of the 12-block Chronos-2 encoder as its frozen numerical front end. The hidden width is 768 and the patch length is 16. The numerical predictor contains self-attention over prediction queries and cross-attention to visible numerical memory. Entity descriptions, event content, and covariate names are represented by cached 4,096-dimensional embeddings from the frozen Qwen3-VL-Embedding-8B encoder \citep{li2026qwen3vlembedding}. Unit-normalized embeddings are multiplied by the square root of their dimension before learned projection. The context encoder uses paired additive fusion: event content and time share one token, and each covariate name and numerical summary share one token.

The covariate summary has 12 entries: transformed mean, standard deviation, first and last values; standardized change and slope; observed fraction and count; and four standardized temporal-bin means. Summaries use only retained historical observations. Prediction queries combine learned target tokens with position, calendar, frequency, lead-time, and task information. The main pretraining task mixture uses forecasting and masked interior-block prediction with equal probability and supervises at most six target patches per sampled task.

\begin{table}[ht]
\centering
\caption{Pythia configurations. Counts are in millions and include retained frozen numerical parameters. The external semantic encoder, training-only reference copy, optimizer states, and anchor buffers are excluded; shared parameters are counted once.}
\label{tab:sizes}
\begin{tabular*}{\linewidth}{@{\extracolsep{\fill}}lrrrrr@{}}
\toprule
Size & Context layers & Predictor layers & Total & Stage 1 trained & Stage 2 trained\\
\midrule
Tiny & 1 & 1 & 149.64 & 30.16 & 28.32\\
Small & 1 & 2 & 159.09 & 39.61 & 28.32\\
Medium & 1 & 3 & 168.54 & 49.07 & 28.32\\
Large & 2 & 4 & 185.08 & 65.60 & 28.32\\
XLarge & 3 & 5 & 201.62 & 82.14 & 28.32\\
\bottomrule
\end{tabular*}
\end{table}

\subsection{Objective evaluation and optimization}

\paragraph{Stable reference computation.}
For Equation~\ref{eq:reference}, let $\Delta=a-b$ and let $\eta$ be the smallest positive normal float32 value. For $b>0$, the implementation evaluates
\begin{equation}
\ell(a,b)=\frac{\Delta+|\Delta|}{2}
+b\log\!\left(1+\exp\!\left(-\frac{|\Delta|}{\max(b,\eta)}\right)\right),
\label{eq:reference-stable}
\end{equation}
using float32 errors and \texttt{log1p}; when $b=0$, it returns $a$. The nonpositive exponential argument avoids overflow. Direction normalization uses $\epsilon_n=10^{-12}$, separately from the $10^{-8}$ log-RMS stabilization.

\paragraph{Optimization.}
We use AdamW with learning rate $10^{-4}$, weight decay $0.01$, momentum coefficients $(0.9,0.999)$, gradient clipping at 1, 2,000 warmup updates, and a cosine schedule over each run's declared update budget. Training uses bfloat16 and seed 42. During Stage~2, decoder blocks 8, 9, and 10 (zero indexed) are adapted with relative learning rates $1,0.1,0.01$. Block 11, the final normalization, and the native quantile head remain frozen. The parameter anchor coefficient is $10^{-3}$; the penalty averages the mean squared displacement from pretrained initialization within each trainable tensor and then averages over tensors.

The headline Tiny configuration uses eight devices with batch size 128 per device and 10,000 updates per stage. Its selected Stage~2 checkpoint is at update 10,000, and it consumes 10.24 million sampled windows in each stage. The Small mechanism reference uses the same batch and 30,000 updates per stage, selecting Stage~2 update 30,000. Medium uses batch 256 and 50,000 updates per stage, with Stage~1 supplying its final checkpoint and Stage~2 selected at update 45,000. Large uses batch 128 and 10,000 updates per stage, selecting update 10,000. Each model uses one checkpoint across all benchmark tasks. Sampling is online; Appendix~\ref{app:budget} defines the corresponding exposure counts.

\subsection{Training protocols for the learning-recipe controls}
Direct forecasting skips Stage~1 and updates the context encoder, predictors, and permitted decoder parameters using the Stage~2 forecasting objective for 60K updates. Joint readout adaptation reuses the reference Stage~1 checkpoint and updates the representation along with the readout for 30K Stage~2 updates. Numerical decoder adaptation has no Pythia context/world branch and adapts native numerical blocks 8, 9, and 10 for 60K updates using the same relative rates and anchor. Its front end, last block, final normalization, and quantile head remain frozen. The numerical-only two-stage control instead retains 30K latent-pretraining and 30K readout updates, disabling context and reference guidance throughout training. It shares Pythia's numerical initialization, task mixture, latent-alignment objective, readout recipe, batch size, and checkpoint-selection rule. All controls train on the same pretraining corpus, with no MUSE-specific optimization.

\section{Pretraining corpus}
\label{app:data}

The pretraining corpus combines public time series with data generated internally by Ant International. Following the MUSE domain taxonomy, we report this nonpublic portion as \emph{Industrial}, without disclosing constituent sources or finer categories. Table~\ref{tab:corpus} lists the public source families represented in the eligible window pools; a family name does not imply that all of its subsets are used. MUSE held-out evaluation examples are deliberately excluded from pretraining. For benchmark-linked sources, training uses only admissible observations preceding the held-out cutoffs.

\input{sections/corpus_table}

\begin{figure}[ht]
\centering
\includegraphics[width=\linewidth]{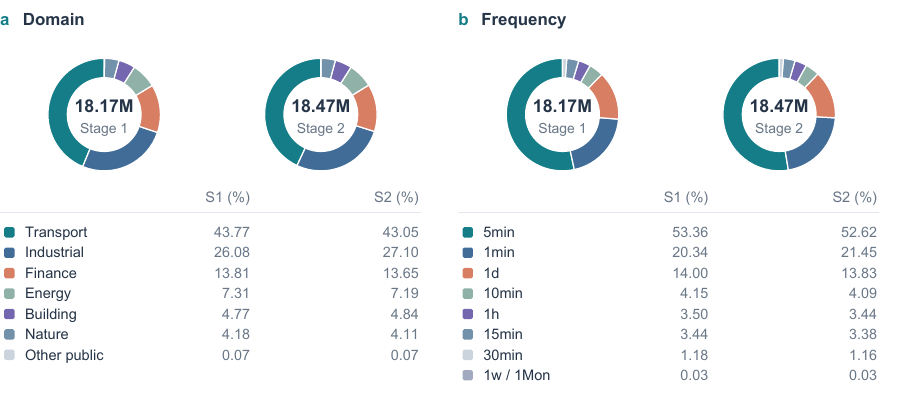}
\caption{Composition of the eligible pretraining window pools by domain and observation frequency, shown separately for the two stages. S1/S2 denote Stage~1/Stage~2; min, h, d, w, and Mon denote minute, hour, day, week, and month. Shares count unique eligible history origins after exclusions and validation removal, not repeated training draws. Internal sources are combined as Industrial. Small public-domain categories are grouped for readability.}
\label{fig:corpus}
\end{figure}

Stage~1 contains \textbf{18,173,268} eligible windows and Stage~2 contains \textbf{18,474,132}. We count each unique frequency--entity--history-origin tuple once after exclusions, temporal cutoffs, eligibility filtering, and validation-entity removal; an origin eligible under multiple tasks or horizons is not counted repeatedly. The stages have different eligibility pools and are counted separately. Figure~\ref{fig:corpus} shows their domain and frequency composition, with Transport, Industrial, and Finance as the largest domains. These corpus proportions differ from the sampling distribution used during optimization.

Training covers nine frequencies: 1 minute, 5 minutes, 10 minutes, 15 minutes, 30 minutes, 1 hour, 1 day, 1 week, and 1 month. Five-minute observations account for 53.36\%/52.62\% of the two eligible pools, with additional sampling support for scarce low-frequency series. Maximum numerical histories are 2,880 observations for intraday/hourly series, 1,440 for daily series, 360 for weekly series, and 180 for monthly series.

\section{MUSE evaluation protocol}
\label{app:protocol}

\subsection{Metrics and aggregation}
We use the MUSE interface and its normalized MASE and WSQL. MASE uses the forecast median and a seasonal scale computed from observed context. WSQL uses deciles $\mathcal Q=\{0.1,0.2,\ldots,0.9\}$, with twice the pinball loss normalized by absolute target magnitude. These quantiles approximate an integrated probabilistic score; the reported WSQL is not an exact continuous ranked probability score (CRPS). Both metrics retain MUSE's validity masks and coverage rules.

Let $\mathcal G$ denote the set of dataset--frequency--horizon scoring groups. Within group $g\in\mathcal G$, the evaluator first pools the appropriate sufficient statistics over valid targets to obtain $m_g$. It divides the result by the Seasonal Naive score $b_g$ on matching coverage. The overall score $S$ is
\begin{equation}
S=\exp\!\left(\frac{1}{|\mathcal G|}\sum_{g\in\mathcal G}\log\frac{m_g}{b_g}\right),
\qquad |\mathcal G|=29.
\end{equation}
This equal-group geometric mean differs from averaging individual windows or the 14 dataset scores. Seasonal Naive has normalized score 1. Missing probabilistic scores for point-only baselines remain missing. Horizon summaries in Figure~\ref{fig:groups} apply the same geometric mean to the corresponding subset of original groups.

\subsection{Evaluation coverage and model comparisons}
Our evaluation retains the last at most 20 rolling windows for each public entity--horizon configuration and all prescribed Industrial windows. This gives 661 entity--horizon configurations and 12,729 test windows across 29 scoring groups. Target windows do not overlap within a configuration. All 31 baseline rows in Table~\ref{tab:main} come directly from MUSE Table~3, preserving the published scores, training regimes, and missing probabilistic results.

Main and method-comparison results use the checkpoint selected on the held-out pretraining subset; capacity and budget slices use each independently budgeted run's final checkpoint. Differences between these same-seed runs are reported descriptively, without claims of statistical significance.

\input{sections/context_coverage}

\section{Detailed forecasting results}
\label{app:groups}

\subsection{Results by scoring group, horizon, and frequency}

\begin{figure}[ht]
\centering
\includegraphics[width=\linewidth]{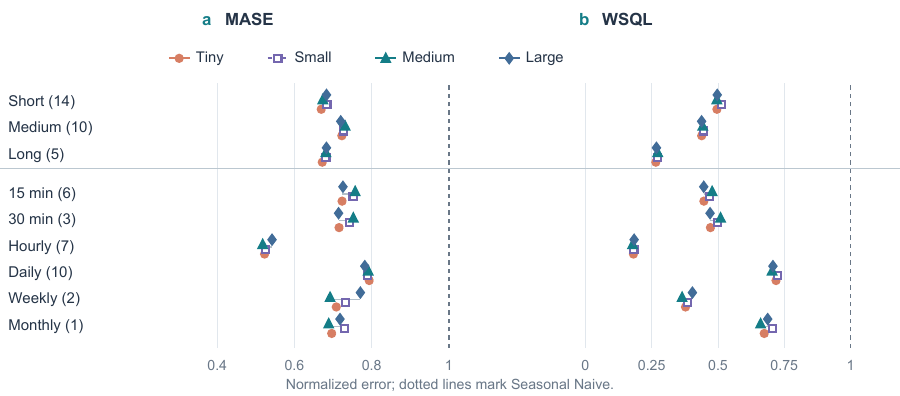}
\caption{MUSE performance across forecast horizons and observation frequencies. Each point is the equal-weight geometric mean of its original normalized scoring groups; parentheses give group counts. Vertical dotted lines mark Seasonal Naive at 1. The same model checkpoint is used throughout. These slices have different task mixtures, so their scores do not isolate horizon or frequency effects on identical data.}
\label{fig:groups}
\end{figure}

Table~\ref{tab:groups} reports all scoring groups for the four main-table Pythia configurations; their geometric means recover the overall results. Figure~\ref{fig:groups} summarizes these scores by horizon and frequency. CiK abbreviates ServiceNow's Context is Key datasets.

\input{sections/group_table}

\subsection{Forecast examples}

\begin{figure}[ht]
\centering
\includegraphics[width=\linewidth]{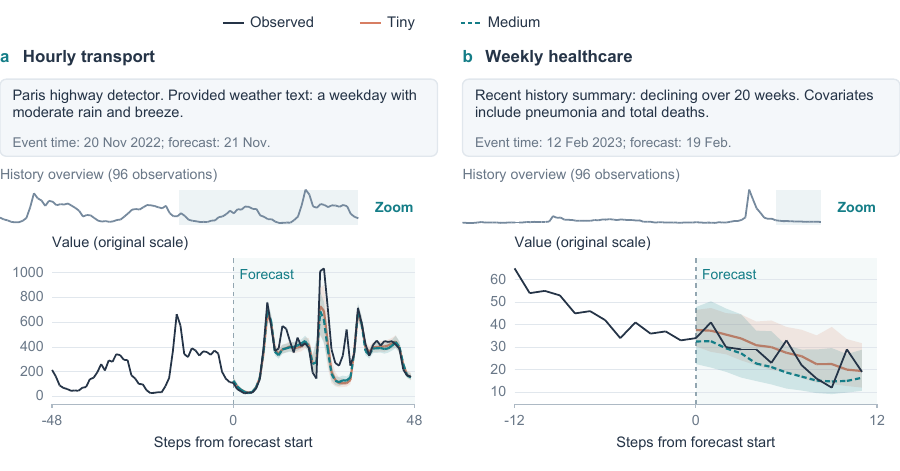}
\caption{Two public MUSE forecast windows with Tiny and Medium predictions: hourly transport data from CGTSF-PTF and weekly healthcare data from TimeCAP. Curves show predictive medians and bands span the 0.1--0.9 quantiles, in the original observation scale. The overview strips show the final 96 exported history observations; the larger panels show one forecast horizon of history (48 hours or 12 weeks). Context cards summarize benchmark-supplied inputs retained for the prediction origin. The examples use fixed public configurations and windows selected independently of forecast quality.}
\label{fig:examples}
\end{figure}

Figure~\ref{fig:examples} illustrates the shape and spread of Tiny and Medium predictions at two temporal resolutions. Both models use the same numerical history and retained benchmark context. Appendix~\ref{app:quantile-diagnostics} complements these examples with coverage and ordering statistics over the complete test set.

\input{sections/quantile_diagnostics}

\section{Model capacity and training budget}
\label{app:budget}

\input{sections/scaling_analysis}

\section{Representation learning and readout}
\label{app:additional}

\subsection{Training objectives and input controls}

\begin{figure}[t]
\centering
\includegraphics[width=\linewidth]{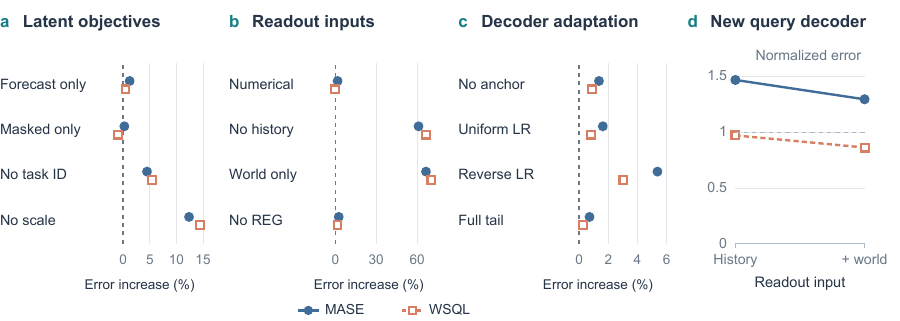}
\caption{Latent representation and readout controls in Small. ID denotes identifier and LR denotes learning rate. (a--c) Percentage error increases relative to Pythia-Small for changes to latent objectives, readout inputs, and decoder adaptation. Axes differ to preserve the scale of the history-removal effect. (d) Final 30K results for a newly trained query decoder using history alone or history plus world states; the dashed horizontal line marks Seasonal Naive. The first three panels use validation-selected checkpoints. Details and absolute scores appear in Table~\ref{tab:additional}.}
\label{fig:latent-readout}
\end{figure}
\input{sections/ablation_table}
\input{sections/additional_ablations}
\input{sections/numerical_initialization}

\input{sections/latent_diagnostics}

\section{Context analysis}
\label{app:context}

\subsection{Test-time semantic interventions}
\input{sections/context_analysis}

\clearpage
\subsection{Training-time context controls by benchmark group}
\begin{figure}[H]
\centering
\includegraphics[width=\linewidth]{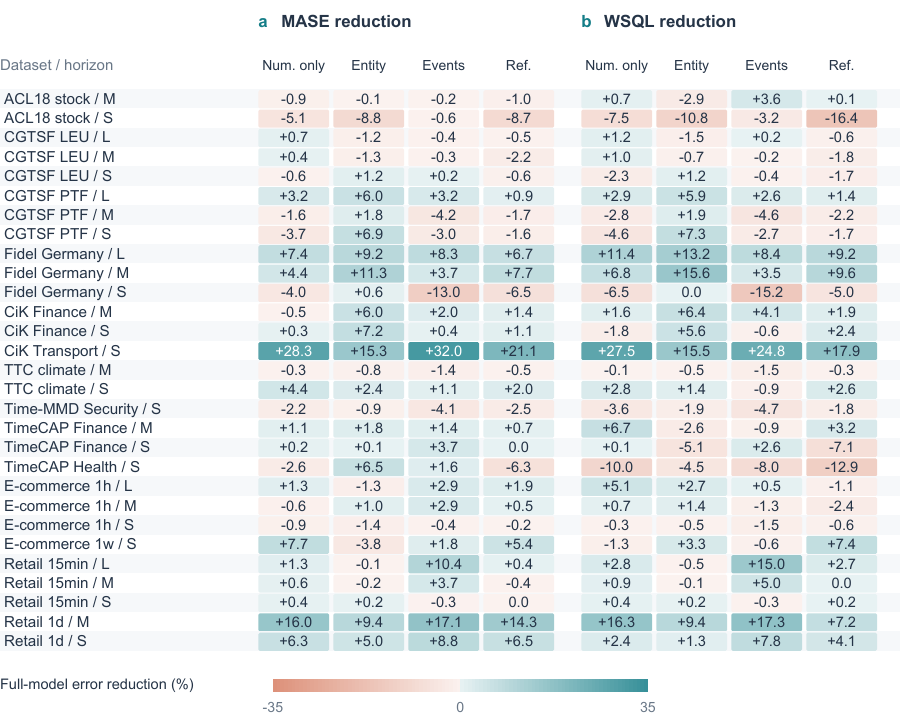}
\caption{Relative error reductions of full Pythia-Small over numerical-only two-stage learning (Num.\ only), no-entity, no-event, and no-reference controls in every MUSE scoring group. Writing $s_{\mathrm{full}}$ and $s_{\mathrm{control}}$ for the corresponding normalized MASE or WSQL scores in that group, the reduction is $100(1-s_{\mathrm{full}}/s_{\mathrm{control}})$. Teal favors the full model and coral favors the control; both panels use the same symmetric color scale. S/M/L denote short/medium/long horizons. All controls use matched Small recipes and selected checkpoints. Negative cells remain visible, showing that aggregate improvements need not hold for every task.}
\label{fig:context-groups}
\end{figure}
\input{sections/numerical_group_table}

The matched numerical-only comparison in Table~\ref{tab:numerical-groups} separates multimodal predictive learning from numerical JEPA pretraining under the same two-stage schedule. Full Pythia reduces MASE/WSQL by 28.28\%/27.51\% on short-horizon CiK transport and 15.96\%/16.32\% on medium-horizon daily retail. All 50 CiK transport windows contain only entity context alongside numerical history. Conversely, full Pythia's errors increase by 5.08\%/7.52\% relative to numerical-only learning on short-horizon ACL18 stock. These separately trained controls measure changes to shared predictive learning, not per-window attribution: removing a source during pretraining can affect groups where that source is absent at inference. The comparison supports aggregate benefits, with task-dependent gains and regressions.

\clearpage
\input{sections/related_work_extended}

%% file: sections/corpus_table.tex
\begin{table}[H]
\centering
\small
\caption{Training source families and unique eligible window counts, after exclusions and training/validation separation. Context-Guided Time Series Forecasting is abbreviated CGTSF. Public family names link to their sources; internal data are disclosed only as Industrial. Inclusion of a family does not imply inclusion of its MUSE evaluation examples. The two stage-specific pools overlap and must not be added as distinct examples.}
\label{tab:corpus}
\begin{tabular*}{\linewidth}{@{\extracolsep{\fill}}lrr@{}}
\toprule
Source family & Stage 1 windows & Stage 2 windows\\
\midrule
\href{https://github.com/yumoxu/stocknet-dataset}{ACL18} & 6,099 & 6,271\\
\href{https://huggingface.co/datasets/ChengsenWang/CGTSF}{CGTSF} & 100,686 & 100,266\\
\href{https://www.kaggle.com/datasets/blacksnail789521/time-imm/data}{EPA-Air} & 2,414 & 2,414\\
\href{https://huggingface.co/datasets/Zihan1004/FNSPID}{FNSPID} & 2,493,885 & 2,505,451\\
\href{https://huggingface.co/collections/wxcai/fidel-ts-68ef3c7813c4187e2c88dedd}{Fidel-TS} & 10,718,993 & 10,743,093\\
\href{https://www.kaggle.com/datasets/blacksnail789521/time-imm/data}{ILINet} & 278 & 279\\
\href{https://github.com/daydreamer-amelia/From_News_to_Forecast}{NewsForecast} & 90,090 & 90,075\\
\href{https://www.kaggle.com/datasets/blacksnail789521/time-imm/data}{RepoHealth} & 3,490 & 3,510\\
\href{https://studentlife.cs.dartmouth.edu/dataset.html}{StudentLife} & 6,916 & 7,104\\
\href{https://github.com/Rose-STL-Lab/Multimodal_Forecasting}{TTC} & 680 & 688\\
\href{https://github.com/AdityaLab/Time-MMD}{Time-MMD} & 3,071 & 3,067\\
\href{https://github.com/geon0325/TimeCAP}{TimeCAP} & 815 & 829\\
\href{https://github.com/haoxin1998/TimesX-project}{TimesX} & 5,477 & 5,435\\
Industrial (internal) & 4,740,374 & 5,005,650\\
\bottomrule
\end{tabular*}
\end{table}

%% file: sections/context_coverage.tex
\subsection{Context availability and preprocessing}
\label{app:context-coverage}

Table~\ref{tab:context-coverage} counts retained inputs for 12,729 MUSE test-window instances across 661 configurations and 29 scoring groups. Availability follows the model's effective masks: a valid entity slot, at least one retained event, or at least one retained covariate with a finite historical observation. The frequency-description token is not an event. Event inputs are truncated at the last observed history timestamp using the supplied timestamps. Covariate summaries use finite observations from retained historical rows only. Entity descriptions are nonempty in every window; events and covariates occur in 73.96\% and 82.53\%, respectively, and 59.92\% contain all three. These are window-level proportions; MUSE scores give equal weight to the 29 groups. ACL18 windows contain entity information and covariates, both CGTSF sources contain entity information and events, and e-commerce windows contain all three.

\input{sections/context_coverage_table}

%% file: sections/context_coverage_table.tex
\begin{table}[htbp]
\centering
\small
\caption{Observed context availability in MUSE under the normal-input Small preprocessing. Bits denote effective entity/event/covariate inputs in that order. Counts are benchmark window instances; percentages use 12,729 as the denominator. All eight combinations were counted, including the four zero-count patterns.}
\label{tab:context-coverage}
\begin{tabular*}{\linewidth}{@{\extracolsep{\fill}}llrr@{}}
\toprule
Effective inputs & Pattern & Windows & Share\\
\midrule
Entity only & 100 & 436 & 3.43\%\\
Entity + events & 110 & 1,788 & 14.05\%\\
Entity + covariates & 101 & 2,878 & 22.61\%\\
Entity + events + covariates & 111 & 7,627 & 59.92\%\\
Other four combinations & 000--011 & 0 & 0.00\%\\
\midrule
Total & & 12,729 & 100.00\%\\
\bottomrule
\end{tabular*}
\end{table}

%% file: sections/group_table.tex
\begin{table}[htbp]
\centering
\small
\caption{All 29 MUSE scoring groups for the four reported Pythia configurations. Each pair gives normalized MASE / WSQL ($\downarrow$); Seasonal Naive is 1 in every group. Freq. denotes observation frequency and Term denotes forecast horizon; S/M/L denote short/medium/long horizons. CGTSF subsets LEU and PTF denote London Electricity Usage and Paris Traffic Flow, respectively. The selected checkpoint is shared across all tasks.}
\label{tab:groups}
\begin{tabular*}{\linewidth}{@{\extracolsep{\fill}}lllrrrr@{}}
\toprule
Dataset & Freq. & Term & Tiny & Small & Medium & Large\\
\midrule
ACL18 stock & 1d & M & 0.989/1.007 & 0.999/0.998 & 0.971/0.934 & 0.982/1.014\\
ACL18 stock & 1d & S & 1.014/0.959 & 1.133/1.168 & 1.014/0.949 & 1.018/0.933\\
CGTSF\_LEU & 30min & L & 0.711/0.398 & 0.741/0.421 & 0.741/0.427 & 0.711/0.397\\
CGTSF\_LEU & 30min & M & 0.717/0.437 & 0.747/0.462 & 0.745/0.465 & 0.712/0.433\\
CGTSF\_LEU & 30min & S & 0.718/0.603 & 0.737/0.635 & 0.771/0.666 & 0.720/0.604\\
CGTSF\_PTF & 1h & L & 0.511/0.290 & 0.544/0.311 & 0.544/0.315 & 0.540/0.307\\
CGTSF\_PTF & 1h & M & 0.520/0.317 & 0.575/0.355 & 0.581/0.360 & 0.550/0.335\\
CGTSF\_PTF & 1h & S & 0.349/0.301 & 0.425/0.372 & 0.432/0.372 & 0.357/0.307\\
Fidel-TS Germany & 15min & L & 0.866/0.333 & 0.919/0.366 & 0.964/0.395 & 0.884/0.343\\
Fidel-TS Germany & 15min & M & 0.881/0.359 & 0.979/0.410 & 0.980/0.411 & 0.897/0.362\\
Fidel-TS Germany & 15min & S & 0.570/0.413 & 0.617/0.458 & 0.594/0.446 & 0.562/0.405\\
CiK Finance & 1d & M & 0.750/0.488 & 0.762/0.492 & 0.780/0.503 & 0.769/0.500\\
CiK Finance & 1d & S & 0.688/0.633 & 0.687/0.647 & 0.695/0.648 & 0.700/0.637\\
CiK Transport & 1h & S & 0.595/0.527 & 0.539/0.454 & 0.565/0.470 & 0.658/0.528\\
TTC\_climate & 1d & M & 0.561/0.456 & 0.571/0.466 & 0.567/0.464 & 0.558/0.454\\
TTC\_climate & 1d & S & 0.650/0.643 & 0.649/0.653 & 0.668/0.656 & 0.653/0.643\\
Time-MMD Security & 1Mon & S & 0.697/0.674 & 0.730/0.705 & 0.689/0.661 & 0.718/0.687\\
TimeCAP finance & 1d & M & 0.946/0.880 & 0.933/0.880 & 0.882/0.829 & 0.942/0.907\\
TimeCAP finance & 1d & S & 1.001/0.969 & 1.005/0.955 & 0.945/0.867 & 0.968/0.931\\
TimeCAP healthcare & 1w & S & 0.549/0.443 & 0.559/0.476 & 0.522/0.455 & 0.621/0.499\\
e\_commerce\_1H & 1h & L & 0.596/0.063 & 0.553/0.059 & 0.516/0.052 & 0.604/0.062\\
e\_commerce\_1H & 1h & M & 0.570/0.070 & 0.522/0.065 & 0.485/0.059 & 0.576/0.069\\
e\_commerce\_1H & 1h & S & 0.568/0.099 & 0.540/0.097 & 0.516/0.093 & 0.565/0.098\\
e\_commerce\_1W & 1w & S & 0.914/0.322 & 0.958/0.314 & 0.919/0.292 & 0.956/0.327\\
retail\_15T & 15min & L & 0.731/0.545 & 0.719/0.519 & 0.731/0.544 & 0.724/0.530\\
retail\_15T & 15min & M & 0.671/0.530 & 0.673/0.530 & 0.682/0.540 & 0.674/0.530\\
retail\_15T & 15min & S & 0.671/0.559 & 0.672/0.561 & 0.674/0.561 & 0.671/0.558\\
retail\_1D & 1d & M & 0.786/0.722 & 0.681/0.650 & 0.821/0.733 & 0.691/0.648\\
retail\_1D & 1d & S & 0.706/0.685 & 0.660/0.659 & 0.693/0.646 & 0.694/0.651\\
\bottomrule
\end{tabular*}
\end{table}

%% file: sections/quantile_diagnostics.tex
\subsection{Quantile coverage and ordering}
\label{app:quantile-diagnostics}

\begin{figure}[ht]
\centering
\includegraphics[width=\linewidth]{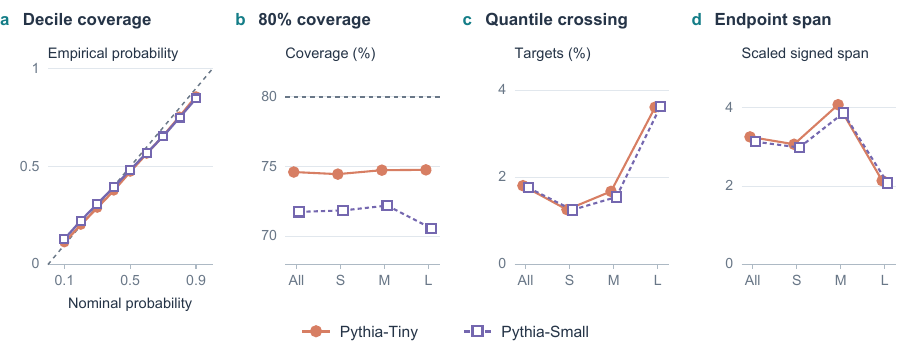}
\caption{Quantile diagnostics under normal inputs. (a) Empirical decile coverage against nominal probability. (b) Central-80\% coverage, with its nominal target. (c) The proportion of targets with any adjacent crossing. (d) Mean signed endpoint span normalized by the MASE scale. Overall and horizon-specific summaries average the original scoring groups equally: 29 overall, 14 short, 10 medium, and five long.}
\label{fig:quantile-diagnostics}
\end{figure}

The selected Tiny and Small checkpoints each provide finite predictions at all nine native deciles for 1,189,817 valid targets across 12,729 normal-input MUSE windows. Let $y$ denote an observed target and $\hat x^{(\tau)}$ its predicted quantile at level $\tau$. For each scoring group, we compute the empirical fraction $\Pr(y\leq\hat x^{(\tau)})$, central-80\% coverage, and the fraction of targets with any adjacent quantile crossing. With $s$ denoting the positive seasonal-error scale used by MASE, we also measure the signed endpoint span $(\hat x^{(0.9)}-\hat x^{(0.1)})/s$. Figure~\ref{fig:quantile-diagnostics} and Table~\ref{tab:quantile-diagnostics} report equal-weight arithmetic averages over the corresponding scoring groups, using the original quantile outputs without reordering or calibration.

\begin{table}[ht]
\centering
\small
\caption{Group-macro quantile diagnostics. The short, medium, and long subsets contain 253,250, 464,260, and 472,307 valid target points, respectively. Signed span retains any endpoint inversions instead of repairing the predictions.}
\label{tab:quantile-diagnostics}
\begin{tabular*}{\linewidth}{@{\extracolsep{\fill}}llrrr@{}}
\toprule
Model & Horizon & 80\% coverage (\%) & Any crossing (\%) & Scaled signed span\\
\midrule
Pythia-Tiny & All & 74.62 & 1.81 & 3.260\\
 & Short & 74.47 & 1.26 & 3.072\\
 & Medium & 74.75 & 1.68 & 4.082\\
 & Long & 74.77 & 3.62 & 2.142\\
\midrule
Pythia-Small & All & 71.77 & 1.77 & 3.137\\
 & Short & 71.87 & 1.26 & 2.989\\
 & Medium & 72.22 & 1.55 & 3.871\\
 & Long & 70.60 & 3.63 & 2.086\\
\bottomrule
\end{tabular*}
\end{table}

Tiny's central-80\% coverage exceeds Small's by 2.85 percentage points and accompanies a larger scaled signed span. Coverage stays near 74.5--74.8\% across Tiny's horizon groups and ranges from 70.6\% to 72.2\% for Small, below the nominal 80\% level for both models. Crossing rates are higher at long horizons and concentrated in Fidel-TS Germany, where they range from 10.96\% to 15.01\% for Tiny and 11.52\% to 15.12\% for Small. Pooling target points across groups gives 36,894/1,189,817 (3.10\%) crossings for Tiny and 37,497/1,189,817 (3.15\%) for Small. These summaries complement WSQL with the coverage and ordering of the native quantile outputs.

%% file: sections/scaling_analysis.tex
\subsection{Capacity and sampled training exposure}

The sweep varies the context encoder and predictor over a shared numerical initialization and frozen front end (Table~\ref{tab:sizes}). Stage~1 trains 30.16--82.14M parameters, while Stage~2 adapts 28.32M throughout. Retained model sizes include frozen numerical parameters and exclude the external semantic encoder.

Let $U_1$ and $U_2$ denote optimizer updates in representation pretraining and readout training, with total budget $U=U_1+U_2$. The eight-device experiments use the same per-device batch size $b$ in both stages, without gradient accumulation. Let $D_1,D_2$ denote the sampled-window exposures in the two stages and $D$ their total:
\begin{equation}
D_1=8bU_1,\qquad D_2=8bU_2,\qquad D=D_1+D_2=8bU.
\end{equation}
Exposure counts repeated draws, distinct from the unique eligible windows in Appendix~\ref{app:data}. Equal $U$ or $D$ does not imply equal computation: the stages update different modules, and larger configurations change the cost of an update.

\input{sections/scaling_table}

\paragraph{Capacity and update budget.}
Figure~\ref{fig:scaling} and Table~\ref{tab:scaling} report the complete five-size sweep at batch 128 and 10K, 30K, or 50K updates per stage. The configuration with the lowest error on both metrics shifts from Tiny at 10K (0.6879/0.4269), to Small at 30K (0.6981/0.4373), and to Medium at 50K (0.7103/0.4393), although the optimum itself does not improve monotonically. Medium benefits from extending the paired budgets from 10K to 30K to 50K in both metrics; Tiny and XLarge do not. The sweep thus reveals an interaction between capacity and optimization budget: increasing either alone does not reliably improve performance. These local results do not establish a scaling law or exhausted model capacity.

\subsection{Batch size and stage allocation}

\begin{figure}[t]
\centering
\includegraphics[width=\linewidth]{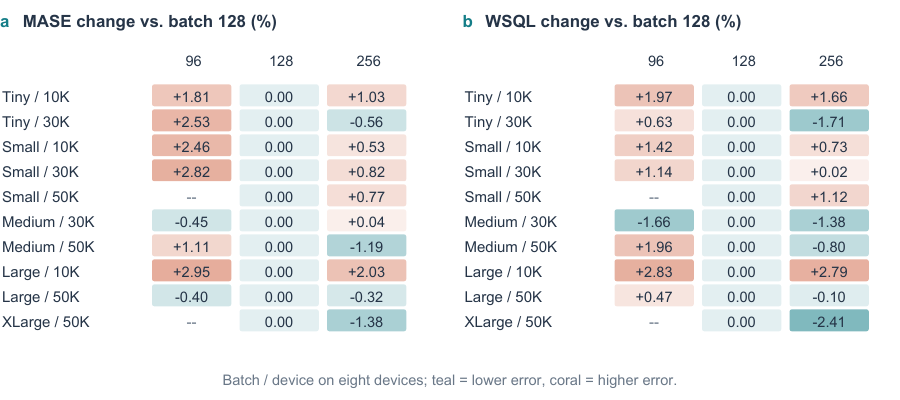}
\caption{All ten measured batch slices, using final checkpoints. Let $s_b$ denote the MASE or WSQL score at per-device batch size $b$. Cells show percentage changes from batch 128 within the same size and per-stage update budget: $100(s_b/s_{128}-1)$. Teal denotes improvement and coral degradation, using a common symmetric scale of $\pm4\%$. Dashes mark unmeasured settings. Both metrics exhibit interactions with size and duration; larger batches are not uniformly better. Absolute scores are in Table~\ref{tab:batch}.}
\label{fig:batch-interactions}
\end{figure}

\input{sections/batch_table}

\paragraph{Batch size.}
The complete batch slices in Figure~\ref{fig:batch-interactions} and Table~\ref{tab:batch} make these interactions explicit. For Tiny at 10K, batch 128 is preferable to both 96 and 256; at 30K, increasing batch from 96 to 128 to 256 improves both metrics. Medium shows an even clearer reversal: batch 96 has the lowest MASE and WSQL at 30K, whereas at 50K both improve monotonically with batch, from 0.7182/0.4479 to 0.7103/0.4393 to 0.7019/0.4357. Small behaves differently: batch 256 does not improve either metric over 128 at any of its measured 10K, 30K, and 50K budgets. For Large at 50K, WSQL improves across the three batches while MASE is nonmonotonic. The preferred batch therefore depends on capacity and duration; changing batch at fixed updates also changes sampled exposure.

\begin{figure}[t]
\centering
\includegraphics[width=\linewidth]{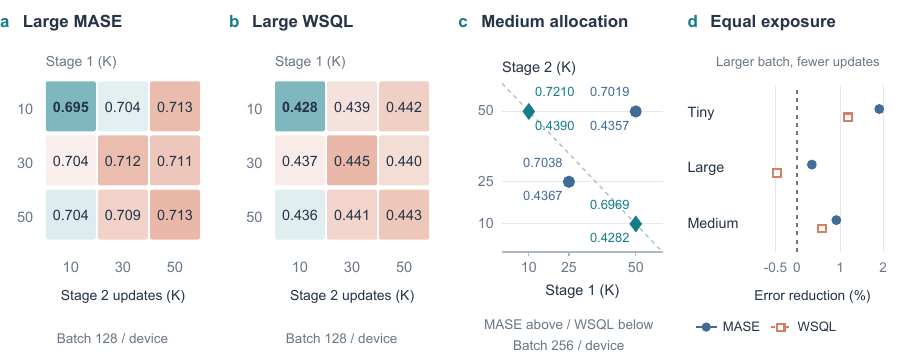}
\caption{Allocation and sampled exposure at final checkpoints. (a,b) The full Large Stage~1$\times$Stage~2 budget grid, batch 128; each panel's shading ranges from its lowest error (teal) to highest (coral). (c) Medium at batch 256, with MASE above and WSQL below each point; diamonds on the dashed diagonal share 60K total updates. (d) Equal-exposure exchanges between larger batches and more updates: Tiny/Large compare 10K+10K at batch 256 with 20K+20K at batch 128; Medium compares 25K+25K at 256 with 50K+50K at 128. Positive values favor the larger-batch recipe.}
\label{fig:allocation}
\end{figure}

\paragraph{Stage allocation.}
Figure~\ref{fig:allocation} includes the full Large grid with $U_1,U_2\in\{10,30,50\}$K at batch 128. For every Stage~1 budget, the 10K readout has lower MASE and WSQL than the 30K and 50K readouts. Along the fixed-$60$K total-update comparison, 50K+10K obtains 0.7039/0.4363, compared with 0.7127/0.4417 for 10K+50K and 0.7125/0.4453 for 30K+30K. However, the best point in the complete grid is 10K+10K (0.6955/0.4279), so the evidence favors neither increasing total updates unconditionally nor assigning all additional updates to Stage~1.

The Medium comparisons at batch 256 show a related allocation effect. At 60K total updates, 50K+10K obtains 0.6969/0.4282 versus 0.7210/0.4390 for 10K+50K, reducing MASE by 3.34\% and WSQL by 2.46\%. Using the same 50K representation checkpoint with an independently scheduled 50K readout yields 0.7019/0.4357 at the final checkpoint. A balanced 25K+25K allocation obtains 0.7038/0.4367. Together, the two sizes show that readout duration is a substantive part of the training recipe rather than a neutral continuation of representation learning.

\paragraph{Matched sampled exposure.}
Three comparisons preserve exposure separately in each stage while exchanging batch size for update count. Tiny with 10K+10K updates at batch 256 and 20K+20K at batch 128 uses 20.48 million draws per stage in both cases. The shorter, larger-batch recipe obtains 0.6950/0.4339 instead of 0.7085/0.4392, improving MASE/WSQL by 1.91\%/1.19\%. For Medium, 25K+25K at batch 256 and 50K+50K at batch 128 both use 51.20 million draws per stage; the former improves the two metrics by 0.91\%/0.58\%. Large supplies a useful counterpoint: the same 10K--20K exchange as Tiny improves MASE slightly (0.7096 versus 0.7121) but worsens WSQL (0.4398 versus 0.4378). Exposure alone therefore does not determine the outcome. These comparisons also change the number of optimizer updates and the schedule horizon, and should be interpreted as alternative recipes with matched sampled exposure.

\subsection{Checkpoint trajectories}

\begin{figure}[t]
\centering
\includegraphics[width=\linewidth]{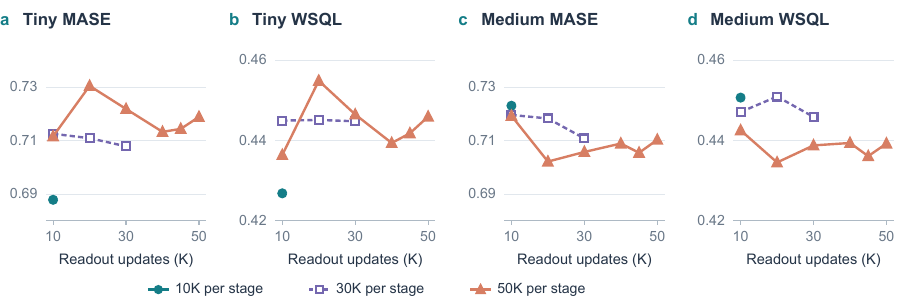}
\caption{MUSE test trajectories within Tiny and Medium runs at batch 128 per device. Colors identify independent pretraining budgets of 10K, 30K, or 50K updates in each stage; points within a curve are saved Stage~2 checkpoints from that run. Each curve therefore begins from its own pretrained representation and follows its own readout schedule. These curves describe transfer during training; checkpoint selection uses the separate pretraining validation subset.}
\label{fig:checkpoint-dynamics}
\end{figure}

The checkpoint trajectories in Figure~\ref{fig:checkpoint-dynamics} answer a different question from the final-checkpoint matrix. Within Tiny's 30K+30K, batch-128 run, readout MASE improves from 0.7126 at 10K to 0.7079 at 30K, yet its final score remains above that of the independently trained 10K+10K model. The longer recipe changes both the representation-training budget and the readout learning-rate horizon; it is not a continuation of the short recipe. Medium's 50K+50K, batch-128 trajectory is also nonmonotonic: readout MASE is 0.7022 at 20K and 0.7103 at its final 50K endpoint. The observed reversals alone cannot identify overfitting or establish insufficient data or parameter scale as their cause. They instead delimit the regime in which Pythia currently exhibits reliable gains and motivate treating its two training stages jointly when reasoning about scale.

%% file: sections/scaling_table.tex
\begin{table}[H]
\centering
\small
\caption{Capacity and budget results at final checkpoints. Updates are in thousands; batch size is per device on eight devices. Each run uses its own declared learning-rate horizon. More updates or larger batches change sampled exposure, including repetitions, rather than the number of unique training windows.}
\label{tab:scaling}
\begin{tabular*}{\linewidth}{@{\extracolsep{\fill}}lrrrr@{}}
\toprule
Size & Stage 1 + Stage 2 (K) & Batch & MASE & WSQL\\
\midrule
Tiny & 10 + 10 & 128 & 0.6879 & 0.4269\\
Tiny & 30 + 30 & 128 & 0.7079 & 0.4448\\
Tiny & 50 + 50 & 128 & 0.7189 & 0.4460\\
Small & 10 + 10 & 128 & 0.6999 & 0.4344\\
Small & 30 + 30 & 128 & 0.6981 & 0.4373\\
Small & 50 + 50 & 128 & 0.7152 & 0.4486\\
Medium & 10 + 10 & 128 & 0.7231 & 0.4507\\
Medium & 30 + 30 & 128 & 0.7108 & 0.4459\\
Medium & 50 + 50 & 128 & 0.7103 & 0.4393\\
Large & 10 + 10 & 128 & 0.6955 & 0.4279\\
Large & 30 + 30 & 128 & 0.7125 & 0.4453\\
Large & 50 + 50 & 128 & 0.7132 & 0.4429\\
XLarge & 10 + 10 & 128 & 0.7130 & 0.4431\\
XLarge & 30 + 30 & 128 & 0.7207 & 0.4497\\
XLarge & 50 + 50 & 128 & 0.7340 & 0.4557\\
\midrule
\multicolumn{5}{l}{\emph{Medium: stage allocation and batch size}}\\
Medium & 10 + 50 & 256 & 0.7210 & 0.4390\\
Medium & 50 + 10 & 256 & 0.6969 & 0.4282\\
Medium & 25 + 25 & 256 & 0.7038 & 0.4367\\
Medium & 50 + 50 & 256 & 0.7019 & 0.4357\\
Medium & 50 + 50 & 96 & 0.7182 & 0.4479\\
\bottomrule
\end{tabular*}
\end{table}

%% file: sections/batch_table.tex
\begin{table}[htbp]
\centering
\small
\caption{All measured batch-size slices at final checkpoints. Entries are MASE / WSQL; $b$ is batch size per device on eight devices. Stage~1 and Stage~2 each use the indicated update budget. Dashes indicate unmeasured combinations.}
\label{tab:batch}
\begin{tabular*}{\linewidth}{@{\extracolsep{\fill}}lrrrr@{}}
\toprule
Size & Updates/stage (K) & $b=96$ & $b=128$ & $b=256$\\
\midrule
Tiny & 10 & 0.7003/0.4353 & 0.6879/0.4269 & 0.6950/0.4339\\
Tiny & 30 & 0.7258/0.4476 & 0.7079/0.4448 & 0.7039/0.4372\\
Small & 10 & 0.7171/0.4406 & 0.6999/0.4344 & 0.7036/0.4376\\
Small & 30 & 0.7178/0.4423 & 0.6981/0.4373 & 0.7038/0.4374\\
Small & 50 & -- & 0.7152/0.4486 & 0.7207/0.4536\\
Medium & 30 & 0.7076/0.4385 & 0.7108/0.4459 & 0.7111/0.4398\\
Medium & 50 & 0.7182/0.4479 & 0.7103/0.4393 & 0.7019/0.4357\\
Large & 10 & 0.7160/0.4399 & 0.6955/0.4279 & 0.7096/0.4398\\
Large & 50 & 0.7104/0.4450 & 0.7132/0.4429 & 0.7110/0.4425\\
XLarge & 50 & -- & 0.7340/0.4557 & 0.7239/0.4447\\
\bottomrule
\end{tabular*}
\end{table}

%% file: sections/ablation_table.tex
\begin{table}[t]
\centering
\small
\caption{Small-model comparisons on MUSE. Pythia, numerical-only two-stage, and context controls use 30K updates per stage; direct forecasting and numerical decoder adaptation use 60K updates in one stage. The numerical-only model omits context and reference guidance throughout both stages. Joint readout adaptation reuses the same Stage~1 and updates the representation during Stage~2. All rows use validation-selected checkpoints. $\Delta$ is relative MASE change from Pythia-Small.}
\label{tab:ablation}
\begin{tabular*}{\linewidth}{@{\extracolsep{\fill}}lrrr@{}}
\toprule
Variant & MASE $\downarrow$ & WSQL $\downarrow$ & $\Delta$ MASE\\
\midrule
\multicolumn{4}{l}{\emph{Learning recipe}}\\
Pythia-Small & 0.6981 & 0.4373 & +0.00\%\\
Direct forecasting & 0.7174 & 0.4442 & +2.77\%\\
Joint readout adaptation & 0.7836 & 0.4942 & +12.25\%\\
Numerical decoder adaptation & 0.7148 & 0.4419 & +2.39\%\\
Numerical-only two-stage & 0.7148 & 0.4464 & +2.40\%\\
\midrule
\multicolumn{4}{l}{\emph{Context and reference guidance}}\\
Without entity information & 0.7168 & 0.4475 & +2.68\%\\
Without event information & 0.7197 & 0.4462 & +3.10\%\\
Without covariates & 0.7097 & 0.4460 & +1.67\%\\
Without reference guidance & 0.7086 & 0.4404 & +1.51\%\\
\bottomrule
\end{tabular*}
\end{table}

%% file: sections/additional_ablations.tex
\begin{table}[ht]
\centering
\small
\caption{Additional Small comparisons with normal inputs, global batch 1,024, and 30K updates per stage. Results use the validation-selected Stage~2 checkpoint: 25K for the history-only query readout and 30K otherwise. The query readouts share the pretrained representation and use a separately trained two-layer query decoder.}
\label{tab:additional}
\begin{tabular*}{\linewidth}{@{\extracolsep{\fill}}lrr@{}}
\toprule
Variant & MASE $\downarrow$ & WSQL $\downarrow$\\
\midrule
Pythia-Small & 0.6981 & 0.4373\\
\midrule
\multicolumn{3}{l}{\emph{Pretraining tasks and latent scale}}\\
Forecasting task only & 0.7066 & 0.4392\\
Masked reconstruction only & 0.6997 & 0.4333\\
Without task embedding & 0.7294 & 0.4609\\
Without log-RMS alignment & 0.7845 & 0.5002\\
\midrule
\multicolumn{3}{l}{\emph{Context representation}}\\
Without context embedding normalization & 0.7242 & 0.4423\\
Separate instead of paired context tokens & 0.7094 & 0.4438\\
\midrule
\multicolumn{3}{l}{\emph{Readout inputs and adaptation}}\\
Numerical predictions instead of world states & 0.7099 & 0.4370\\
Without history in readout & 1.1245 & 0.7275\\
World states only in readout & 1.1627 & 0.7455\\
Without REG token & 0.7157 & 0.4440\\
Without decoder weight anchor & 0.7076 & 0.4412\\
Uniform decoder learning rates & 0.7094 & 0.4409\\
Reversed decoder learning rates & 0.7355 & 0.4506\\
Adapt all tail blocks, normalization, and head & 0.7030 & 0.4384\\
\midrule
\multicolumn{3}{l}{\emph{Alternative query readout}}\\
History alone & 1.4170 & 0.9353\\
History and predicted world states & 1.2948 & 0.8627\\
\bottomrule
\end{tabular*}
\end{table}

\paragraph{Pretraining tasks and latent scale.}
Figure~\ref{fig:latent-readout} organizes the comparisons around the information learned by the representation and the information retained by the readout. Forecast-only pretraining is weaker than the mixed task on both metrics. Masked-only pretraining is close in MASE (0.6997 versus 0.6981) and better in WSQL (0.4333 versus 0.4373), so the mixture offers a point-forecasting advantage without uniformly dominating probabilistic accuracy. Removing task conditioning worsens MASE/WSQL by 4.48\%/5.39\%, supporting an explicit distinction between forecasting and masked reconstruction. Removing log-RMS alignment has a larger effect, increasing the two scores by 12.38\%/14.38\%. This is consistent with latent magnitude carrying useful predictive information beyond direction; the error comparison does not directly measure the geometry of the learned states.

\paragraph{Context representation.}
Removing unit normalization of semantic embeddings increases MASE to 0.7242 and WSQL to 0.4423. Replacing paired additive context tokens with separate tokens also worsens both metrics, to 0.7094/0.4438. These controls indicate that passing more tokens or retaining the raw embedding magnitude does not automatically improve the use of context. In the present architecture, aligning input scales and coupling content to its temporal or numerical attributes provide a more effective interface to the predictive state. The field-removal controls in Table~\ref{tab:context-reference} assess the contribution of the information itself.

\input{sections/context_reference_table}

\begin{samepage}
\paragraph{Context combinations and reference guidance.}
Table~\ref{tab:context-reference} holds the two-stage recipe fixed while varying context and reference guidance. Without guidance, entity/event text improves both metrics over numerical-only; covariates improve WSQL but slightly worsen MASE. All context gives the lowest errors among no-reference variants, and guidance improves both further. Table~\ref{tab:reference-objectives} compares additional choices for the reference objective.
\par\end{samepage}

\begin{table}[ht]
\centering
\small
\caption{Alternative reference objectives under the same Small training recipe and validation-selected 30K readout checkpoint. All rows retain numerical initialization, contextual inputs, latent matching, and log-RMS alignment.}
\label{tab:reference-objectives}
\begin{tabular*}{\linewidth}{@{\extracolsep{\fill}}lrr@{}}
\toprule
Auxiliary objective & MASE $\downarrow$ & WSQL $\downarrow$\\
\midrule
Numerical-reference guidance (default) & 0.6981 & 0.4373\\
No reference objective & 0.7086 & 0.4404\\
Fixed-temperature numerical reference & 0.7127 & 0.4421\\
Numerical/null/random reference & 0.7047 & 0.4365\\
\bottomrule
\end{tabular*}
\end{table}

\paragraph{Alternative reference objectives.}
The fixed-temperature variant averages errors within each sample before applying a unit-temperature softplus to the corrected-minus-detached-numerical error. The numerical/null/random variant instead uses the minimum of the numerical, null-context, and random-context errors as its reference. The default objective adapts its scale to the numerical error of each patch before aggregation. It gives the lowest MASE among these variants; the numerical/null/random reference gives slightly lower WSQL (0.4365 versus 0.4373). The fixed-temperature variant is weaker on both metrics. These comparisons connect reference construction and aggregation to forecasting quality, alongside the latent-state measurements in Appendix~\ref{app:latent-diagnostics}.

\subsection{Readout inputs and decoder design}
\label{app:readout-design}

\paragraph{Readout inputs.}
The numerical-prediction control retains multimodal Stage~1 but trains and evaluates the readout using numerical predictions in place of corrected world states. This substitution increases MASE but slightly lowers WSQL. Removing historical numerical memory is much more damaging, whether the native REG token is retained (1.1245/0.7275) or removed as well (1.1627/0.7455). By comparison, removing REG while retaining history yields 0.7157/0.4440. The readout benefits from their complementary roles: history provides observed numerical memory, and world states contribute a context-conditioned prediction of the future.

\paragraph{Alternative query decoder.}
The query-readout pair keeps the same pretrained representation and changes whether a newly trained two-layer query decoder can attend to world predictions in addition to numerical history. Table~\ref{tab:additional} reports the validation-selected checkpoints, which differ in update count. Figure~\ref{fig:latent-readout} additionally compares the two final 30K checkpoints: adding world states reduces MASE from 1.4683 to 1.2948 and WSQL from 0.9711 to 0.8627, improvements of 11.81\% and 11.15\%. Both variants remain substantially weaker than the adapted native readout and even Seasonal Naive on MASE. The gains show that the predicted states contain useful information for the query decoder; recovering the main model's observation-space accuracy also depends on the decoder interface.

%% file: sections/context_reference_table.tex
\begin{table}[htbp]
\centering
\small
\caption{Context combinations and numerical-reference guidance under the Small 30K+30K protocol. All models are pretrained with their listed context inputs and share the task mixture, numerical initialization, readout policy, batch size, and validation selection rule. These are training controls, not test-time missing-modality interventions.}
\label{tab:context-reference}
\begin{tabular*}{\linewidth}{@{\extracolsep{\fill}}llrr@{}}
\toprule
Context & Reference guidance & MASE $\downarrow$ & WSQL $\downarrow$\\
\midrule
None (numerical only) & No & 0.7148 & 0.4464\\
Entity + events & No & 0.7134 & 0.4439\\
Covariates & No & 0.7168 & 0.4449\\
All context & No & 0.7086 & 0.4404\\
All context & Yes & 0.6981 & 0.4373\\
\bottomrule
\end{tabular*}
\end{table}

%% file: sections/numerical_initialization.tex
\subsection{Numerical initialization and adaptation policies}
\label{app:initialization}

\paragraph{Adaptation policies and training protocol.}
Figure~\ref{fig:numerical-initialization} and Table~\ref{tab:initialization} examine all nine combinations of frozen (F), anchored adaptation (A), and unanchored adaptation (U) policies. Every configuration initializes its numerical components from Chronos-2. The first policy controls all eight numerical encoder blocks during Stage~1; the second controls all four decoder blocks during Stage~2. A and U train the same blocks at $0.1$ times the main learning rate, with anchor coefficients $10^{-3}$ and zero, respectively. The anchor penalizes displacement from pretrained initialization, averaging squared displacement within each trainable tensor and then over tensors. Tokenization, REG embeddings, final normalization, and the quantile head remain frozen. The target encoder tracks only the numerical encoder by EMA, with decay increasing from 0.996 to 0.9999. It remains fixed under F and evolves under A/U.

All configurations use Small, global batch 1,024, and 30K Stage~1 updates. Each row shares one Stage~1 endpoint; its decoder branches start independently from the original numerical tail. A/U columns train for 30K updates. F columns perform zero Stage~2 updates and have no validation-selected checkpoint. During Stage~2, the encoder and all predictive/contextual representation modules are frozen, including encoders adapted in Stage~1. The representation is therefore fixed before the decoder policy is applied. The matrix's F/A cell also differs from default Pythia: the latter adapts three decoder blocks with descending learning-rate multipliers $1,0.1,0.01$, rather than four blocks at a uniform $0.1$.

\input{sections/initialization_table}

\paragraph{Encoder preservation and decoder adaptation.}
The frozen-encoder row has the lowest errors under every decoder policy. However, retaining an unadapted decoder is insufficient: F/F reaches 1.3271 MASE, compared with 0.7031 for F/A. Anchored decoder adaptation reduces MASE by 47.02\%, 59.35\%, and 57.72\% relative to the frozen decoder in the F, A, and U rows, respectively. Adaptation therefore matters even when the numerical encoder is preserved. This pattern is consistent with the need to align the native decoder with predicted multimodal states. The poorer adapted-encoder results may reflect encoder drift, changing EMA targets, and optimization; the matrix does not isolate these mechanisms or establish catastrophic forgetting.

\paragraph{Anchoring across stages.}
With an adapted decoder, the anchored encoder improves over its unanchored counterpart by 5.83--5.85\% in MASE and 7.41--7.43\% in WSQL, but remains worse than the frozen encoder. This advantage is conditional: with a frozen decoder, A/F improves MASE over U/F by 2.06\% but worsens WSQL by 3.71\%. Anchoring does not uniformly preserve compatibility with the original readout.

Decoder anchoring has much smaller effects in the uniform-rate matrix. F/U selects update 25K, giving 0.7079/0.4369 MASE/WSQL; its 30K endpoint is 0.7032/0.4339, nearly matching F/A's 0.7031/0.4337. All other trained cells select 30K. At matched 30K endpoints, decoder anchoring reduces MASE by only 0.011\%, 0.045\%, and 0.030\% in the F, A, and U rows. Selected-checkpoint differences alone would overstate its influence in the F row. Conversely, removing the anchor from the default descending-rate decoder increases MASE/WSQL by 1.36\%/0.88\% (Table~\ref{tab:additional}). Its usefulness depends on the adaptation policy.

\paragraph{Decoder adaptation scope and learning rates.}
For the default three trainable decoder blocks, equalizing learning-rate multipliers to $0.37,0.37,0.37$ increases MASE/WSQL by 1.62\%/0.82\%; reversing them to $0.01,0.1,1$ increases errors by 5.37\%/3.03\%. Equal multiplier sums do not imply equal parameter updates. Full-tail adaptation retains the original three rates and additionally trains the fourth block, final normalization, and quantile head at multiplier $0.01$. It yields 0.7030/0.4384, slightly worse than default Small's 0.6981/0.4373. This combined change does not isolate individual components. Meanwhile, the matrix's F/A policy worsens MASE by 0.72\% but improves WSQL by 0.84\% relative to default Small: no tested policy dominates every metric. A separate Large uniform-rate experiment obtained 0.7200/0.4441 and is not pooled with the Small comparisons.

\paragraph{Related transfer controls.}
Matched native numerical adaptation for 60K updates reaches 0.7148/0.4419, while jointly updating the pretrained JEPA/context representation during Stage~2 reaches 0.7836/0.4942 (Table~\ref{tab:ablation}). Neither unfreezes the numerical encoder. The readout comparisons in Appendix~\ref{app:readout-design} further assess the use of predicted states. The query decoders are independently and randomly initialized, and their architecture differs from the native tail; neither they nor the policy matrix isolate pretrained versus random initialization in an otherwise identical numerical model. Together, these experiments favor preserving the numerical representation during latent pretraining and selectively adapting its observation-space readout.

%% file: sections/initialization_table.tex
\begin{table}[htbp]
\centering
\small
\caption{Complete numerical-transfer grid in Small. F: frozen; A: trainable with a parameter anchor; U: trainable without an anchor. The first letter controls numerical front blocks 0--7 in Stage~1; the second controls decoder blocks 8--11 in Stage~2. All adapted blocks use learning-rate multiplier 0.1. The front end is frozen throughout Stage~2, and decoder normalization and the quantile head remain frozen in all nine settings. Stage~1 uses 30K updates; F decoder columns skip Stage~2 and have no validation-selected readout. Other columns train for 30K, with the selected readout step shown. Each score pair is MASE / WSQL.}
\label{tab:initialization}
\begin{tabular*}{\linewidth}{@{\extracolsep{\fill}}ccrrr@{}}
\toprule
Front & Decoder & Selected step (K) & Selected / zero-update & Final\\
\midrule
F & F & 0 & 1.3271/0.9020 & 1.3271/0.9020\\
F & A & 30 & 0.7031/0.4337 & 0.7031/0.4337\\
F & U & 25 & 0.7079/0.4369 & 0.7032/0.4339\\
A & F & 0 & 1.9822/1.4013 & 1.9822/1.4013\\
A & A & 30 & 0.8057/0.5046 & 0.8057/0.5046\\
A & U & 30 & 0.8061/0.5047 & 0.8061/0.5047\\
U & F & 0 & 2.0239/1.3512 & 2.0239/1.3512\\
U & A & 30 & 0.8558/0.5451 & 0.8558/0.5451\\
U & U & 30 & 0.8560/0.5451 & 0.8560/0.5451\\
\bottomrule
\end{tabular*}
\end{table}

%% file: sections/latent_diagnostics.tex
\subsection{Latent correction and reference weights}
\label{app:latent-diagnostics}

\begin{figure}[ht]
\centering
\includegraphics[width=\linewidth]{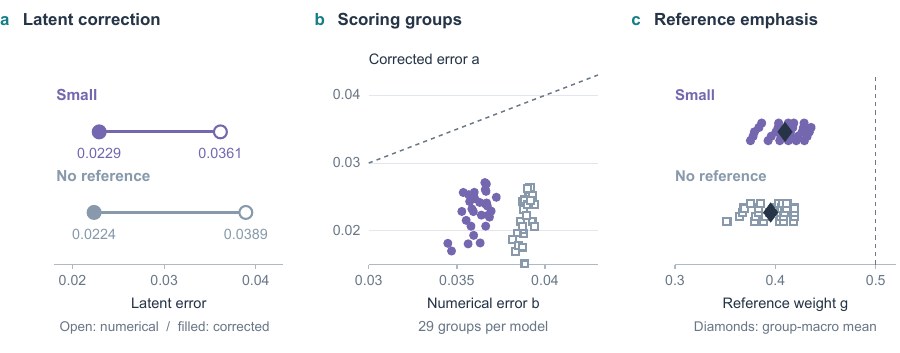}
\caption{Latent correction at fixed Small checkpoints. (a) Group-macro numerical and corrected-state errors. (b) Each point is a scoring group's window-averaged pair $(b,a)$; points below the identity line improve after contextual correction. (c) Distribution of the 29 group-mean reference weights $g=\operatorname{sigmoid}((a-b)/b)$. For the no-reference variant, $g$ is computed as a diagnostic using the same formula.}
\label{fig:latent-diagnostics}
\end{figure}

We evaluate the frozen representations of Pythia-Small and its matched no-reference variant on normal MUSE inputs, using all future query patches and the latent error in Equation~\ref{eq:jepa}. Target states use the same history-derived normalization as prediction. Of 12,729 windows, 12,689 contain valid future patches, yielding 79,102 patches per model; the remaining 40 windows are omitted from latent averages. We first average patch errors within each window, then windows within each scoring group, and finally the 29 groups with equal weight. Figure~\ref{fig:latent-diagnostics} and Table~\ref{tab:latent-diagnostics} report the resulting numerical error $b$ and corrected-state error $a$.

\begin{table}[ht]
\centering
\small
\caption{Latent errors and forecasting results for the matched Small models. Let $\bar a$ and $\bar b$ denote arithmetic group-macro corrected and numerical errors, respectively. Latent reduction is $100(1-\bar a/\bar b)$. MASE and WSQL retain the benchmark's geometric aggregation.}
\label{tab:latent-diagnostics}
\begin{tabular*}{\linewidth}{@{\extracolsep{\fill}}lrrrrr@{}}
\toprule
Model & Numerical $\bar b$ & Corrected $\bar a$ & Reduction & MASE & WSQL\\
\midrule
Pythia-Small & 0.03614 & 0.02291 & 36.59\% & 0.6981 & 0.4373\\
Without reference objective & 0.03891 & 0.02236 & 42.54\% & 0.7086 & 0.4404\\
\bottomrule
\end{tabular*}
\end{table}

Contextual correction lowers mean latent error in all 29 groups for both models: $a<b$ for 79,097 of 79,102 patches (99.994\%) with reference guidance and 79,089 (99.984\%) without it. The guided model has a 7.13\% lower numerical error, with lower $b$ in every group, whereas the no-reference model achieves a larger relative correction and a slightly lower final latent error. Thus, the guided model combines a stronger numerical estimate with better observation-space forecasts, despite the larger latent correction of the no-reference model.

The group-macro reference weights are 0.410 for Pythia-Small and 0.395 for the no-reference diagnostic, with group-specific median patch weights spanning 0.374--0.436 and 0.350--0.421, respectively. All errors are finite and neither model has $b<10^{-6}$: patch-level $b$ ranges from 0.0311 to 0.0415 and from 0.0356 to 0.0422. These evaluations therefore exercise the smooth positive-reference regime of Equation~\ref{eq:reference}.

%% file: sections/context_analysis.tex
\begin{table}[ht]
\centering
\small
\caption{Selected test-time context interventions (MASE). Each row uses the same checkpoint in all three columns. These diagnostics do not affect checkpoint selection.}
\label{tab:context}
\begin{tabular*}{\linewidth}{@{\extracolsep{\fill}}lrrr@{}}
\toprule
Model & Normal & Null & Random\\
\midrule
Pythia-Tiny & 0.6879 & 0.7027 & 0.7315\\
Pythia-Medium & 0.6943 & 0.6966 & 0.6961\\
Pythia-Small & 0.6981 & 0.7089 & 0.7160\\
Small without reference objective & 0.7086 & 0.7063 & 0.7077\\
Small with masked reconstruction only & 0.6997 & 0.6990 & 0.7004\\
\bottomrule
\end{tabular*}
\end{table}

The interventions replace entity, event-content, and covariate-name embeddings with zeros (null) or normalized random vectors (random). Numerical history and numerical covariate summaries remain available. Tiny and Small incur higher MASE under both replacements, whereas the selected Medium checkpoint changes only slightly. The no-reference Small model has worse normal-input forecasts and does not degrade further under replacement; masked-only pretraining also exhibits low sensitivity despite competitive forecasting. Semantic sensitivity thus varies across training recipes and complements the normal-input results and training-time context removals when assessing contextual forecasting utility.

\begin{figure}[ht]
\centering
\includegraphics[width=\linewidth]{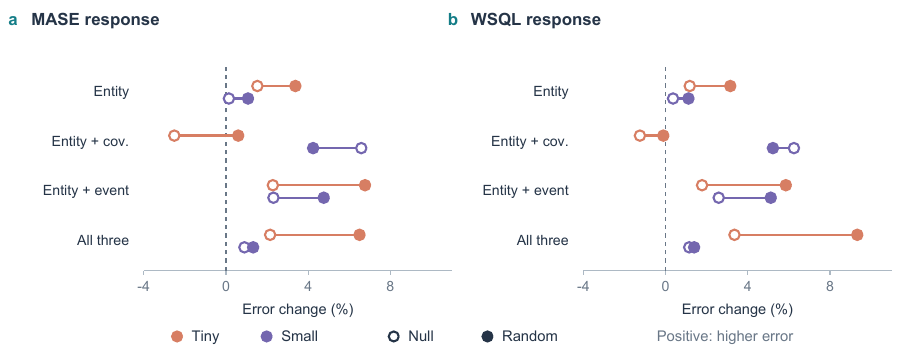}
\caption{Fixed-model responses to joint semantic-embedding replacement, stratified by the context available before intervention; cov. abbreviates covariates. Open and filled markers denote null and random embeddings; colors distinguish Tiny and Small. Within each group--availability slice, $s$ denotes pooled MASE or WSQL, with subscripts indicating intervened/normal inputs; $\operatorname{GM}$ denotes the geometric mean. Each value is $100(\operatorname{GM}(s_{\mathrm{intervened}}/s_{\mathrm{normal}})-1)$ over nonempty scoring-group slices with that availability pattern. The four patterns contain 5/3/11/15 group slices and 436/2,878/1,788/7,627 windows, respectively; groups can contribute to more than one pattern. Numerical histories, event-time attributes, and covariate summaries retain their original values.}
\label{fig:context-sensitivity}
\end{figure}

To locate these responses, we partition the 29 scoring groups by effective input masks into the 34 nonempty group--availability slices in Figure~\ref{fig:context-sensitivity}. Metrics are pooled within the same slice before taking the intervention-to-normal ratio, and ratios are aggregated geometrically across groups.

Tiny's random-embedding WSQL rises by 9.33\% when all three sources are available and by 5.85\% with entity and event inputs; Small's corresponding increases are 1.39\% and 5.12\%. For entity-plus-covariate slices, null embeddings reduce Tiny's MASE by 2.51\% but increase Small's by 6.58\%. The patterns span different task mixtures, including only three groups with entity-plus-covariate inputs, and show that learned semantic responses vary across model configurations and forecasting tasks.

%% file: sections/numerical_group_table.tex
\begin{table}[htbp]
\centering
\small
\caption{Paired Small results for numerical-only two-stage learning and full Pythia. Both use the selected 30K readout checkpoint after 30K latent-pretraining updates, with matching coverage in every group. Score pairs are normalized MASE / WSQL; reduction pairs are percentages, positive when full Pythia improves. These are training-time controls, not test-time removal of context.}
\label{tab:numerical-groups}
\begin{tabular*}{\linewidth}{@{\extracolsep{\fill}}lllrrr@{}}
\toprule
Dataset & Freq. & Term & Numerical only & Full Pythia & Reduction (\%)\\
\midrule
ACL18 stock & 1d & M & 0.9903/1.0053 & 0.9994/0.9984 & -0.92/+0.68\\
ACL18 stock & 1d & S & 1.0783/1.0861 & 1.1331/1.1678 & -5.08/-7.52\\
CGTSF\_LEU & 30min & L & 0.7460/0.4264 & 0.7410/0.4215 & +0.68/+1.17\\
CGTSF\_LEU & 30min & M & 0.7503/0.4664 & 0.7475/0.4618 & +0.39/+0.98\\
CGTSF\_LEU & 30min & S & 0.7329/0.6207 & 0.7370/0.6351 & -0.56/-2.33\\
CGTSF\_PTF & 1h & L & 0.5612/0.3204 & 0.5435/0.3110 & +3.16/+2.93\\
CGTSF\_PTF & 1h & M & 0.5658/0.3459 & 0.5750/0.3554 & -1.62/-2.76\\
CGTSF\_PTF & 1h & S & 0.4098/0.3556 & 0.4248/0.3718 & -3.66/-4.57\\
Fidel-TS Germany & 15min & L & 0.9925/0.4127 & 0.9193/0.3656 & +7.38/+11.41\\
Fidel-TS Germany & 15min & M & 1.0236/0.4394 & 0.9791/0.4097 & +4.35/+6.77\\
Fidel-TS Germany & 15min & S & 0.5927/0.4298 & 0.6166/0.4578 & -4.03/-6.52\\
CiK Finance & 1d & M & 0.7585/0.4995 & 0.7623/0.4917 & -0.50/+1.55\\
CiK Finance & 1d & S & 0.6889/0.6360 & 0.6872/0.6474 & +0.25/-1.78\\
CiK Transport & 1h & S & 0.7513/0.6259 & 0.5388/0.4537 & +28.28/+27.51\\
TTC\_climate & 1d & M & 0.5698/0.4661 & 0.5713/0.4664 & -0.27/-0.05\\
TTC\_climate & 1d & S & 0.6788/0.6713 & 0.6487/0.6526 & +4.44/+2.78\\
Time-MMD Security & 1Mon & S & 0.7142/0.6804 & 0.7298/0.7048 & -2.19/-3.59\\
TimeCAP finance & 1d & M & 0.9427/0.9432 & 0.9328/0.8800 & +1.06/+6.70\\
TimeCAP finance & 1d & S & 1.0067/0.9559 & 1.0046/0.9547 & +0.21/+0.13\\
TimeCAP healthcare & 1w & S & 0.5446/0.4329 & 0.5586/0.4763 & -2.58/-10.02\\
e\_commerce\_1H & 1h & L & 0.5603/0.0617 & 0.5529/0.0585 & +1.31/+5.11\\
e\_commerce\_1H & 1h & M & 0.5191/0.0653 & 0.5221/0.0648 & -0.58/+0.75\\
e\_commerce\_1H & 1h & S & 0.5351/0.0967 & 0.5398/0.0970 & -0.87/-0.27\\
e\_commerce\_1W & 1w & S & 1.0376/0.3095 & 0.9576/0.3135 & +7.70/-1.28\\
retail\_15T & 15min & L & 0.7284/0.5342 & 0.7187/0.5190 & +1.34/+2.83\\
retail\_15T & 15min & M & 0.6773/0.5350 & 0.6733/0.5301 & +0.58/+0.93\\
retail\_15T & 15min & S & 0.6746/0.5629 & 0.6721/0.5606 & +0.37/+0.42\\
retail\_1D & 1d & M & 0.8107/0.7765 & 0.6813/0.6498 & +15.96/+16.32\\
retail\_1D & 1d & S & 0.7047/0.6750 & 0.6602/0.6588 & +6.32/+2.40\\
\bottomrule
\end{tabular*}
\end{table}

%% file: sections/related_work_extended.tex
\section{Extended related work}
\label{app:related_work}

\subsection{Numerical time-series foundation models}

Large-scale pretraining over heterogeneous numerical series has enabled shared forecasting models to transfer across datasets. Chronos tokenizes numerical values for language-model-style probabilistic forecasting \citep{ansari2024chronos}; TimesFM uses a decoder-only architecture \citep{das2024timesfm}; and Moirai accommodates variation in frequency, channel count, and predictive distributions through a universal forecasting architecture \citep{woo2024moirai}. Timer studies generative pretraining across forecasting, imputation, and anomaly detection \citep{liu2024timer}. Time-MoE and Moirai-MoE explore sparse expert architectures for forecasting \citep{shi2025timemoe,liu2025moiraimoe}, while Sundial uses a flow-matching objective to model continuous next-patch distributions without discrete tokenization \citep{liu2025sundial}. Moirai~2.0 combines a decoder-only architecture with quantile forecasting and multi-token prediction \citep{liu2025moirai2}; Timer-S1 and Toto~2 further study scaling numerical forecasting under their respective training recipes \citep{liu2026timers1,khwaja2026toto2}. Chronos-2 extends this line to univariate, multivariate, and covariate-informed forecasting \citep{ansari2025chronos2}.

Other designs emphasize compact architectures or flexible conditioning. Tiny Time Mixers studies lightweight adaptive models for multivariate forecasting \citep{ekambaram2024ttm}. TS-ICL unifies probabilistic forecasting and imputation through timestamp-aligned in-context regression, including covariates and partially observed histories \citep{lenaour2026tsicl}. Pythia builds on the pretrained Chronos-2 encoder and probabilistic decoder, adding context-conditioned prediction within the encoder's latent space.

\subsection{Language models and multimodal forecasting}

\paragraph{Language models for numerical prediction.}
LLMTime serializes numbers as digit strings and performs zero-shot extrapolation through next-token prediction \citep{gruver2023llmtime}. GPT4TS adapts a largely frozen pretrained Transformer across time-series tasks \citep{zhou2023gpt4ts}. Time-LLM reprograms time-series patches through text prototypes and prefix prompts \citep{jin2024timellm}, while AutoTimes maps numerical segments into the language model's embedding space for autoregressive forecasting with timestamp conditioning \citep{liu2024autotimes}. CALF aligns temporal and textual streams at multiple representation levels \citep{liu2025calf}. TimeCMA aligns time-series embeddings with embeddings of prompts derived from the same numerical series \citep{liu2025timecma}.

\paragraph{External context for forecasting.}
GPT4MTS combines textual context with numerical histories for multimodal forecasting \citep{jia2024gpt4mts}. Texts as Time Series converts paired text representations into auxiliary variables usable by numerical forecasters \citep{li2026tats}. VoT combines event reasoning and textual alignment with adaptive fusion of numerical and event predictions \citep{wang2026vot}; TESS extracts structured mean, volatility, shape, and lag information from text for gated contextual integration \citep{li2026tess}. At foundation-model scale, ChatTime supports numerical and textual inputs and outputs in a shared multimodal model \citep{wang2025chattime}, while Aurora targets universal generative multimodal forecasting \citep{wu2026aurora}.

Contextual integration also raises the question of how to train a model to use text effectively. For context-aided forecasting (CAF), DoubleCast addresses the modality gap with semisynthetic CAF-7M training data \citep{zheng2026doublecast}. REST-TS studies text collapse and supervises the textual pathway through structured numerical forecast residuals \citep{nguyen2026restts}. TRACE uses conditional estimation to address missing and irregularly sampled modalities before multimodal fusion \citep{kan2026trace}. Pythia represents entity descriptions, events, and covariate names with a frozen semantic encoder, then uses a trainable context pathway to refine numerical predictive states. Its detached numerical error supplies a reference for contextual learning, with the probabilistic output loss applied in the subsequent readout stage.

\subsection{Forecasting datasets and benchmarks}

Time-MMD provides paired numerical and textual observations across multiple domains \citep{liu2024timemmd}, and Context is Key isolates forecasting problems in which textual information is essential \citep{williams2025cik}. MUSE broadens evaluation to heterogeneous contextual inputs, domains, frequencies, and forecast horizons, comparing numerical foundation models, multimodal methods, and LLM-based forecasters under a common benchmark \citep{chen2026muse}. Numerical benchmarks offer complementary perspectives: the Toto study introduces the Benchmark of Observability Metrics (BOOM) to evaluate forecasting on observability data \citep{cohen2025toto}. We use MUSE as the primary benchmark because contextual observations are central to Pythia's prediction problem. Its published baseline results provide the principal comparison, and we evaluate Pythia on its test tasks without dataset-specific adaptation.

\subsection{JEPA objectives and target representations}

I-JEPA \citep{assran2023ijepa} and V-JEPA \citep{bardes2024vjepa} learn representations by predicting target embeddings rather than reconstructing pixels. VL-JEPA applies predictive learning to vision--language tasks by predicting continuous text embeddings and using a separate decoder to produce text \citep{chen2026vljepa}.

\paragraph{Target spaces and representation stability.}
Variance-invariance-covariance regularization (VICReg) explicitly regularizes embedding variance and covariance \citep{bardes2022vicreg}, and C-JEPA combines it with image-based JEPA to improve representation stability \citep{mo2024cjepa}. SALT demonstrates masked latent prediction with a separately pretrained, frozen teacher in video \citep{li2026salt}. Pythia similarly preserves a pretrained target space in its default recipe, while supervising the direction and magnitude of predicted numerical states. Our encoder/decoder adaptation experiments examine how preserving or updating this initialization affects multimodal predictive learning.

\paragraph{Theoretical perspectives.}
Analyses of deep linear self-distillation identify an implicit preference for predictive features \citep{littwin2024jepabias}. A complementary temporal analysis relates JEPA's emergent dynamical-regime clustering to Koopman-invariant representations and the inductive bias of a near-identity linear predictor \citep{ruizmorales2026koopman}. These studies connect latent prediction to the information retained in learned representations.

\subsection{Temporal JEPA and latent world models}

\paragraph{Latent forecasting and temporal representations.}
LaT-PFN combines a JEPA with a prior-data fitted network (PFN) for zero-shot time-series forecasting \citep{verdenius2024latpfn}. It learns from synthetic series, uses related numerical series as in-context examples, and predicts future embeddings in an EMA-defined target space. Its probabilistic decoder receives detached latent predictions and is optimized with independent gradients. Pythia shares this separation of latent prediction and observation-space decoding, using a pretrained numerical foundation model to define the target space and external multimodal observations to condition its predictive states. It then freezes the predictive modules and adapts a pretrained quantile decoder in a separate training stage.

TS-JEPA predicts masked numerical patch embeddings using an EMA target encoder and evaluates the frozen encoder with classification and regression heads \citep{ennadir2024tsjepa}. CHARM conditions temporal JEPA learning on channel text and transfers frozen embeddings to downstream tasks \citep{dutta2026charm}. SC-JEPA combines a soft codebook bottleneck with prediction at multiple temporal resolutions for anomaly anticipation \citep{he2026scjepa}. Together, these works explore temporal predictive learning, semantic conditioning, and structured target spaces across different downstream tasks.

\paragraph{Latent world models.}
V-JEPA~2 extends predictive learning to large-scale video pretraining and subsequently learns an action-conditioned world model for robotic planning \citep{assran2025vjepa2}. GenRL aligns pretrained vision--language representations with a generative world model for learning embodied behaviors in imagination \citep{mazzaglia2024genrl}. NormWear-2 models physiological signals and clinical interventions through shared latent dynamics \citep{luo2026normwear2}. For multivariate forecasting, Phys-JEPA decomposes states into physical and residual components and constrains latent states and transitions through known physical variables \citep{nie2026physjepa}. Pythia separates a numerical prediction from a correction driven by external context, with both expressed in the pretrained numerical encoder's space.